\RequirePackage{fix-cm}
\documentclass[twocolumn]{svjour3}
\smartqed
\usepackage{graphicx}
\usepackage{cite}
\usepackage{hyperref}
\usepackage{booktabs}
\usepackage{threeparttable}
\usepackage{multirow}
\usepackage{caption}
\usepackage{subcaption}
\usepackage{bm}
\usepackage{bbm}
\usepackage{siunitx}
\usepackage{amssymb}
\usepackage{amsmath}
\usepackage{comment}
\usepackage{calc}
\usepackage{orcidlink}
\usepackage{xspace}
\hypersetup{colorlinks}

\DeclareMathOperator*{\argmin}{arg\,min}
\makeatletter
\DeclareRobustCommand\onedot{\futurelet\@let@token\@onedot}
\def\@onedot{\ifx\@let@token.\else.\null\fi\xspace}

\def\etal{\emph{et~al}\onedot}
\newcommand{\dashrule}[1][black]{
  \color{#1}\rule[\dimexpr.5ex-.2pt]{4pt}{.4pt}\xleaders\hbox{\rule{4pt}{0pt}\rule[\dimexpr.5ex-.2pt]{4pt}{.4pt}}\hfill\kern0pt
}
\makeatother
\begin{document}\sloppy

\title{Design and Analysis of Efficient Attention in Transformers \\for Social Group Activity Recognition}

\author{Masato Tamura\orcidlink{0000-0003-1029-5271}}

\institute{Masato Tamura \at
              Big Data Analytics Solutions Lab, R\&D, \\
              Hitachi America, Ltd., \\
              2535 Augustine Dr, 3rd Floor, Santa Clara, 95054, California, United States. \\
              \email{masato.tamura@ieee.org}
}
\date{Received: date / Accepted: date}
\maketitle

\begin{abstract}
Social group activity recognition is a challenging task extended from group activity recognition, where social groups must be recognized with their activities and group members. Existing methods tackle this task by leveraging region features of individuals following existing group activity recognition methods. However, the effectiveness of region features is susceptible to person localization and variable semantics of individual actions. To overcome these issues, we propose leveraging attention modules in transformers to generate social group features. In this method, multiple embeddings are used to aggregate features for a social group, each of which is assigned to a group member without duplication. Due to this non-duplicated assignment, the number of embeddings must be significant to avoid missing group members and thus renders attention in transformers ineffective. To find optimal attention designs with a large number of embeddings, we explore several design choices of queries for feature aggregation and self-attention modules in transformer decoders. Extensive experimental results show that the proposed method achieves state-of-the-art performance and verify that the proposed attention designs are highly effective on social group activity recognition.
\end{abstract}

\section{Introduction}\label{sec:intro}

\begin{figure*}[t]
    \centering
    \begin{subfigure}[b]{1.0\linewidth}
        \centering
        \includegraphics[width=0.73\linewidth]{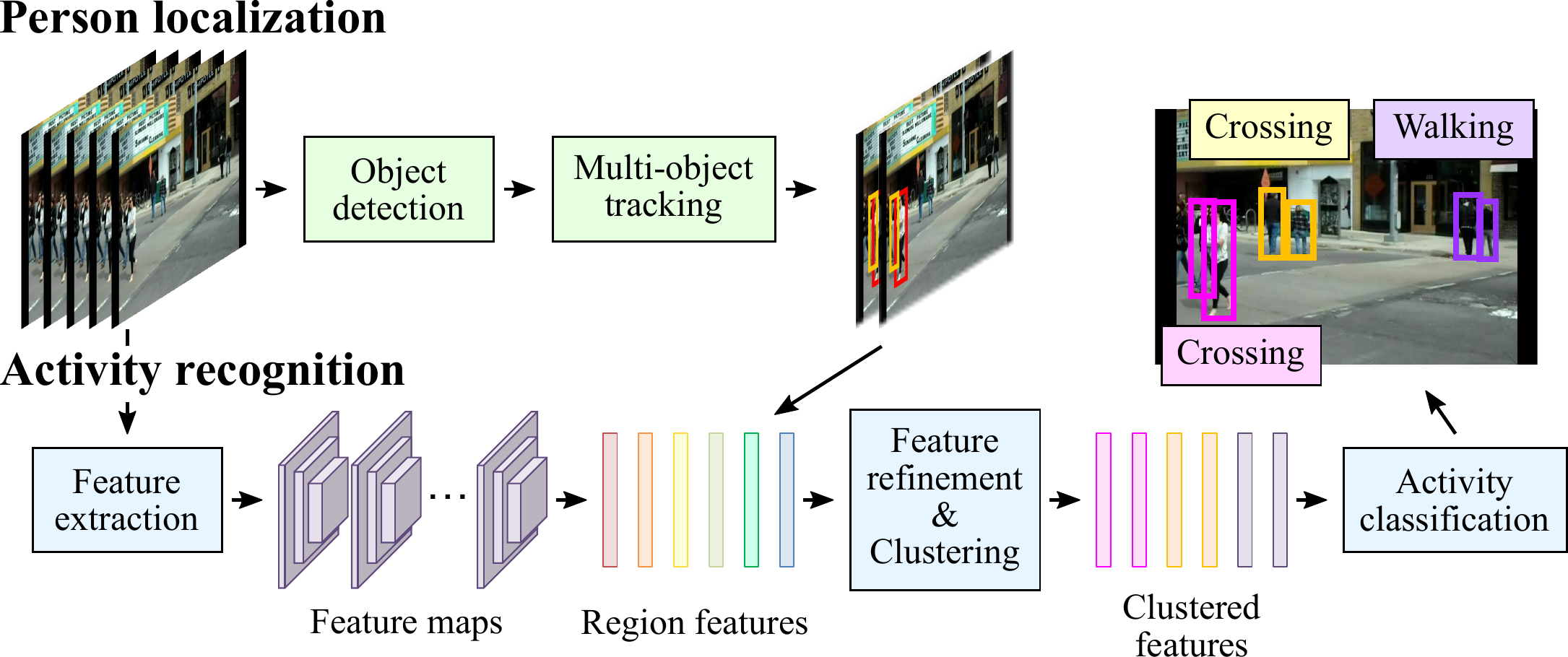}
        \caption{Existing method.}
        \label{fig:intro_exist}
    \end{subfigure} \\
    \begin{subfigure}[b]{1.0\linewidth}
        \centering
        \includegraphics[width=0.73\linewidth]{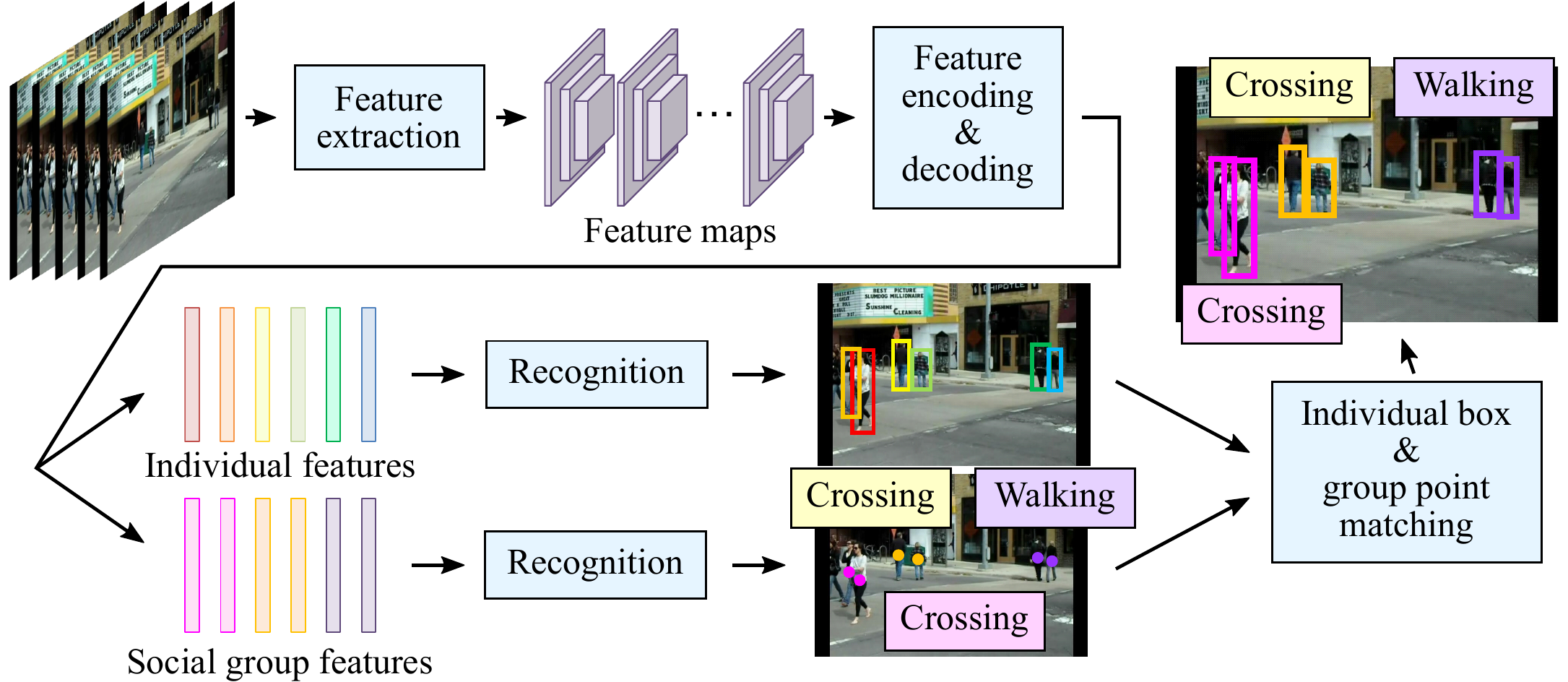}
        \caption{Proposed method.}
        \label{fig:intro_proposed}
    \end{subfigure}
    \caption{Overviews of existing and proposed social group activity recognition methods. Existing methods first extract region features for individuals and then split them into social groups while the proposed method extracts social group features to recognize social group activities and identify group members.}
    \label{fig:exist_proposed}
\end{figure*}

Recognizing humans and their activities in a video sequence is a crucial problem in computer vision with its potential applications in surveillance, sports video analysis, and social scene understanding. Human activities include actions that refer to motions performed by individuals and activities that indicate movements performed by groups of people. Recognizing group activities is typically more challenging than individual actions due to the complex relations of individual motions within a group. Group activity recognition~\cite{wu_cvpr2019,azar_cvpr2019,gavrilyuk_cvpr2020,hu_cvpr2020,ehsanpour_eccv2020,pramono_eccv2020,yan_tpami2020,yan_eccv2020,li_iccv2021,yuan_iccv2021,zhou_arxiv2021,yan_tnnls2021,jinhui_tpami2022} has tackled this challenging task and made considerable progress toward predicting one group activity per scene given the locations of individuals. This task has been recently extended to social group activity recognition~\cite{ehsanpour_eccv2020}, which requires models to recognize social group activities and identify members of each group. Because of this extension, social group activity recognition expands the capability of its applications in real-world scenarios. We focus on developing a social group activity recognition method, which can also be applied to group activity recognition.

Most existing group activity and social group activity recognition methods rely on region features to recognize activities. To utilize region features, the recognition process is roughly divided into two independent steps; person localization and activity recognition. Figure~\ref{fig:intro_exist} shows the recognition process of an existing social group activity recognition method~\cite{ehsanpour_eccv2020}. In this method, people are first localized in frames with bounding boxes using an off-the-shelf object detector and tracker. These boxes are then used to extract region features from feature maps. The extracted region features are refined to encode relationships between individuals and then clustered to split them into social groups. Activities of social groups are finally obtained by classification with clustered region features.

\begin{figure}[t]
    \centering
    \begin{subfigure}[b]{0.4\linewidth}
        \centering
        \includegraphics[width=0.6\linewidth]{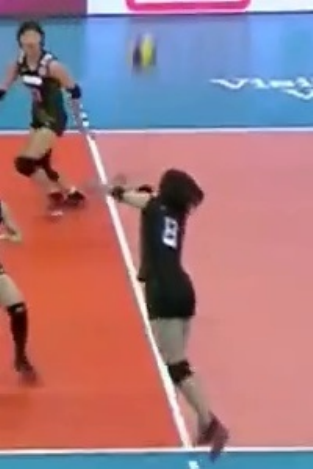}
        \caption{Activity of pass.}
        \label{fig:r_pass_ex_normal}
    \end{subfigure}
    \begin{subfigure}[b]{0.4\linewidth}
        \centering
        \includegraphics[width=0.6\linewidth]{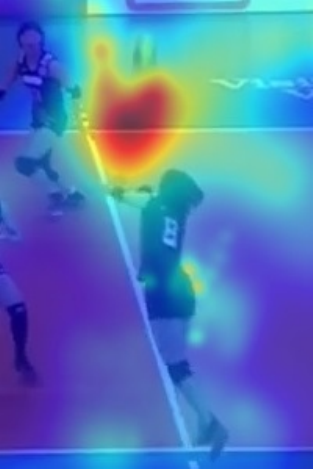}
        \caption{Attention map.}
        \label{fig:r_pass_ex_attn}
    \end{subfigure}
    \caption{Example case where scene contexts are essential for activity recognition. In the Volleyball dataset, the trajectories of balls are one of the important clues for recognizing group activities. The proposed method shows strong attention at the place above players, where balls typically pass.}
    \label{fig:r_pass_ex}
\end{figure}

While region-feature-based methods have significantly improved group activity recognition performance and have successfully been extended to social group activity recognition, they have several drawbacks attributed to the heuristic nature of feature design. Since region features are extracted based on predicted bounding boxes, the effectiveness of region features is subject to localization performance. Existing methods commonly disregard the impact of inaccurate localization during evaluation and use ground truth bounding boxes instead of predicted ones. However, performance degradation has been observed with predicted bounding boxes in several works~\cite{bagautdinov_cvpr2017,qi_eccv2018,wu_cvpr2019,ehsanpour_eccv2020}, which demonstrates the susceptibility of existing methods to person localization. Furthermore, existing methods have limited capability to leverage substantial scene contexts because region features are likely to be dominated by features inside bounding boxes and lack information outside them. Scene contexts, however, sometimes play an essential role in indicating types of group activities. For instance, the trajectories of balls in sports scenes are one of the critical keys for recognizing group activities. Figure~\ref{fig:r_pass_ex_normal} shows the activity of ``Pass" in a volleyball game scene. Typically balls slowly pass above the players who are involved in this activity. This context can be leveraged to distinguish the activity from the intense one such as the activity of ``Spike". Some methods try leveraging scene contexts by pooling a feature map and creating a context embedding, though detailed contexts may not be acquired with this approach.

Existing methods have another issue of region features specifically in social group activity recognition. Since region features are used not only for social group activity recognition but for individual recognition, the performance of group member identification might be affected by the semantics of features for individual recognition. Some group activities occur with group members performing a fixed set of individual actions, while others take place with those doing arbitrary actions. These semantics of individual actions confuse the embedding clustering and as a result, degrade the performance of group member identification.

To address these issues, we previously proposed leveraging a transformer-based object detection framework~\cite{carion_eccv2020,zhu_iclr2021} to generate a social group feature for each social group~\cite{tamura_eccv2022}. The method is designed in such a way that features are aggregated from whole frames with the attention mechanism in transformers. This aggregation renders social group features independent from individual ones as depicted in Fig.~\ref{fig:intro_proposed}. Because of this design, the method has three advantages over existing methods as follows: (1) the effectiveness of social group features is not affected by individual detection, (2) detailed scene contexts can be encoded into social group features, (3) clues for group member identification are aggregated independently from those for individual recognition. Figure~\ref{fig:r_pass_ex_attn} shows the attention values of a transformer in the method. As the figure shows, the method aggregates features not only from the region of the person but the region where balls typically pass, indicating the importance of detailed scene contexts. As a result of this context aggregation and the other advantages, the method achieved state-of-the-art performance on both group activity recognition and social group activity recognition.

Meanwhile, the previous method~\cite{tamura_eccv2022} has limited capability to identify group members especially when group members are far apart from each other or group sizes are large. In our previous design, all the members' locations were encoded into a single feature embedding because group members were identified with their locations, and a single query was used for each social group to aggregate the features of group members. The location of each member was then extracted from the embedding with a feed-forward neural network (FNN). Because of this design, the feature of each member had to be encoded into the different spaces of an embedding. This complex representation of embeddings limited the capability of localization and thus degraded the performance.

This paper extends our previous work~\cite{tamura_eccv2022} to overcome the aforementioned issue. In our extended design, group queries are introduced, each of which is formed of multiple query embeddings. A feature embedding generated with a query embedding is used to localize at most one member in a group, and feature embeddings generated with query embeddings of a group query are combined to predict a group size and activity category. Since features to localize group members are independently encoded into their own embeddings, the same embedding space can be used to encode the location features, simplifying the training of the localization.

It is worth noting that the extension of the previous work~\cite{tamura_eccv2022} is not trivial because the increased number of embeddings renders attention in transformers ineffective. As described in~\cite{carion_eccv2020}, query embeddings are trained to have target localization regions. However, typically a few query embeddings are assigned to ground truth labels per sample during training, and as a result, all the query embeddings may not be trained enough to have target regions. Furthermore, the large number of embeddings renders the self-attention in a transformer decoder ineffective. The naive implementation of self-attention calculates attention values with all the input embeddings and thus hinders detailed communication between embeddings if a large number of embeddings are input. However, detailed communication is essential for the embeddings to identify group members because the communication is done to assign the embeddings of the same group to group members without duplication. To alleviate these problems, we propose efficient designs of group queries and self-attention. In particular, we leverage the insight of the divided attention in spatio-temporal transformers~\cite{bertasius_icml2021} and propose splitting self-attention into iter-group and intra-group self-attention. We compare several design choices including our previous one, the naive implementation, and the proposed one, providing a deep insight into the designs of queries and self-attention for social group activity recognition.

In addition to the new designs, we provide more thorough analyses and extensive experimental results, especially on social group activity recognition. The previous work~\cite{tamura_eccv2022} analyzed the impact of group sizes on recognition performance and revealed the correlation between the performances and the amount of training data. This paper extends the analysis by examining performances in terms of not only group sizes but the distances between group members, revealing the strength of the proposed method in comparison with existing methods and our previous method. Furthermore, the impact of individual actions on the clustering performance in an existing method~\cite{ehsanpour_eccv2020} is analyzed and experimentally verified, which further demonstrates the advantage of the proposed method. Note that these thorough analyses have not been accomplished in the literature on social group activity recognition. We believe these analyses provide new insights, inspire the community, and promote further research in the future.

\section{Related Work}\label{sec:related}

\subsection{Group Activity Recognition}

Group activity recognition was first proposed by Choi~\etal~\cite{choi_iccvw2009} to classify pedestrians' actions on the basis of their collective behaviors with hand-crafted features. Since then, several works have tackled this problem by extracting hand-crafted features of individuals and representing individuals as nodes of probabilistic graphical models~\cite{lan_cvpr2012,lan_tpami2012,amer_iccv2013,amer_eccv2014,amer_tpami2016,wang_cvpr2013}.

With the rise of deep learning to dominate the field of computer vision, several recurrent-neural-network (RNN)-based methods have achieved significant performance due to their learning capability of highly contextualized spatio-temporal relationships~\cite{ibrahim_cvpr2016,bagautdinov_cvpr2017,shu_cvpr2017,wang_cvpr2017,kong_icassp2018,qi_eccv2018}. Ibrahim~\etal~\cite{ibrahim_cvpr2016} proposed architecture to capture the spatial dynamics of people in a scene and the temporal dynamics of each person between frames with RNNs. Bagautdinov~\etal~\cite{bagautdinov_cvpr2017} used a joint learning approach to optimize person localization and activity recognition simultaneously, where RNNs merged and propagated information of region features in the temporal domain. Qi~\etal~\cite{qi_eccv2018} employed a semantic graph to describe the spatial context of the whole scene and incorporated the temporal factor into the graph via RNNs. Some works~\cite{bagautdinov_cvpr2017,qi_eccv2018} tested their models with predicted bounding boxes in addition to ground truth ones, demonstrating the susceptibility of region-feature-based methods to person localization.

To further enhance the capability of capturing spatio-temporal contexts and relationships between people in a scene, graph-neural-network (GNN)-based methods have been proposed~\cite{wu_cvpr2019,ehsanpour_eccv2020,hu_cvpr2020,yuan_iccv2021}. Wu~\etal~\cite{wu_cvpr2019} regarded each individual as a node of relation graphs and applied graph convolutional networks~\cite{kipf_iclr2017} to perform relational reasoning on the graphs. Graph attention networks (GATs)~\cite{velickovic_iclr2018} were introduced to group activity recognition in Ehsanpour~\etal's work~\cite{ehsanpour_eccv2020} to discover underlying interactions between people. This work also proposed social group activity recognition, where adjacency matrices in GATs played an essential role in dividing people into social groups. Yuan~\etal~\cite{yuan_iccv2021} extended GNNs with dynamic node-specific graphs to capture interactions independently for each person.

With the recent advancements in the applications of transformers~\cite{vaswani_nips2017} to computer vision problems, several works introduced transformers into group activity recognition~\cite{gavrilyuk_cvpr2020,li_iccv2021}. Gavrilyuk~\etal~\cite{gavrilyuk_cvpr2020} applied a vanilla transformer encoder to region features to refine them and enhance the semantic representation. Li~\etal~\cite{li_iccv2021} designed spatio-temporal transformers where group representation was augmented by individual representations containing rich spatial context inside frames and temporal dynamics across them.

This work and our previous work~\cite{tamura_eccv2022} also rely on transformers to generate features that have rich spatio-temporal contexts. However, our methods differ from existing methods in that our methods aggregate group features independently from individual ones, while existing methods generate group features based on region features of individuals. Due to the design, our methods enable transformers to aggregate spatio-temporal contexts from whole frames without being affected by individual recognition, resulting in boosting performance.

\subsection{Social Group Detection}

Another literature related to social group activity recognition is social group detection~\cite{weina_tpami2012,choi_eccv2014,park_cvpr2015}, the goal of which is to detect groups of people in a scene without recognizing their activities. Ge~\etal~\cite{weina_tpami2012} proposed clustering trajectories of pedestrians to find groups of people in the surveillance camera footage. Trajectories were generated by a tracking-by-detection paradigm with a hand-crafted feature-based detector and particle filtering tracker. The generated trajectories were clustered with a bottom-up hierarchical clustering approach, where clusters were gradually merged into one based on Hausdorff distance. Choi~\etal~\cite{choi_eccv2014} introduced structured groups that defined the spatial interaction pattern of people. The optimal set of structured groups in a single image was discovered with an energy minimization framework, where an energy function is defined to capture the pattern of each structured group as well as the compatibility between different groups. Park and Shi~\cite{park_cvpr2015} solved a modified set cover problem to detect social groups without inducing overlapping groups. The detected groups were utilized to predict social saliency in the views of first-person cameras, which is the likelihood of joint attention from a social group.

These works provided deep insights into finding social groups from various viewpoints; however, recognizing the activities of social groups along with identifying group members is essential for social scene understanding. Therefore, our works focus on social group activity recognition.

\subsection{Detection Transformer}

The invention of detection transformers~\cite{carion_eccv2020,zhu_iclr2021} is one of the successful cases of introducing transformers to computer vision problems. Carion~\etal~\cite{carion_eccv2020} first proposed using transformers as feature aggregators to generate object features from feature maps. This feature aggregation enables the proposed detector, called DETR, to obviate any heuristic detection points whose features are used to predict object classes and bounding boxes. As a result of this design, an object detection process becomes more straightforward than conventional detectors while demonstrating state-of-the-art performance.

Deformable-DETR~\cite{zhu_iclr2021} enhanced the performance of DETR by replacing standard transformers with deformable ones. One of the issues in DETR was the computational complexity of transformers, which prevented detectors from leveraging high-resolution feature maps. Deformable transformers combined a sparse sampling of the deformable convolution~\cite{dai_iccv2017} and dynamic weighting of standard attention modules, which significantly reduced the computational complexity of transformers. Due to this reduction, multi-scale feature maps were able to be consumed in transformers, and as a result, the detection performance of small objects was improved.

We extend Deformable DETR~\cite{zhu_iclr2021} for social group activity recognition in order to obviate heuristic detection points, which are difficult to define with the arbitrary number of group members, and leverage multi-scale feature maps that can entail both coarse features of group contexts spreading over whole images and fine-grained features of small objects.

\section{Proposed Method}\label{sec:method}

In this section, we introduce a method to extend a transformer-based object detection framework~\cite{carion_eccv2020,zhu_iclr2021} for social group activity recognition. We first explain the overall architecture of the proposed method and show how social group features are generated with the attention mechanism in Sec.~\ref{subsec:overall}. We then describe two design choices of group queries in Sec.~\ref{subsec:des_query}. The self-attention modules in a transformer decoder also have several design choices, which are illustrated in Sec.~\ref{subsec:des_attn}. The loss functions used during training are explained in Sec.~\ref{subsec:loss}, and finally, group member identification in the inference stage is described in Sec.~\ref{subsec:mem_id}.

\subsection{Overall Architecture}\label{subsec:overall}

\begin{figure*}[t]
 \centering
 \includegraphics[width=0.83\linewidth]{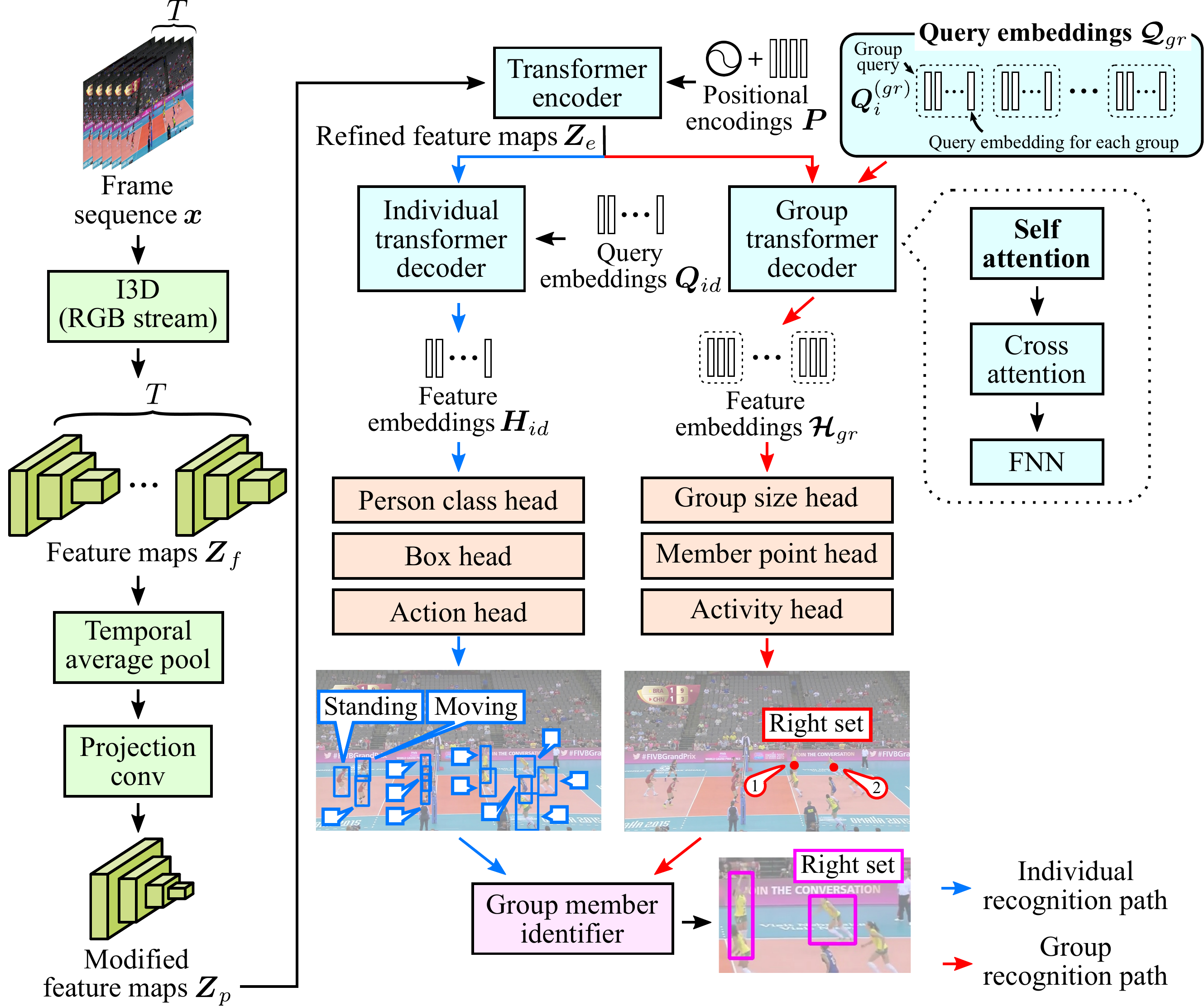}
 \caption{Overall architecture of the proposed method.}\label{fig:overview}
\end{figure*}

Figure~\ref{fig:overview} shows the overall architecture of the proposed method. A frame sequence is first converted to feature maps. The feature maps are then processed through two separate recognition paths; individual recognition path and group recognition path. The individual recognition path generates individual recognition results such as detection scores, bounding boxes, and action scores. In contrast, the group recognition path retrieves group information, namely group sizes, group member locations, and activity scores. Since our focus is on social group activity recognition, the details of the individual recognition process are omitted here and described in the supplementary material of our previous paper~\cite{tamura_eccv2022}.

Given a frame sequence $\bm{x} \in \mathbb{R}^{3 \times T \times H \times W}$, a backbone network extracts a set of multi-scale feature maps $\bm{Z}_{f} = \{\bm{z}_{i}^{\left(f\right)} \mid \bm{z}_{i}^{\left(f\right)} \in \mathbb{R}^{D_{i} \times T \times H_{i}^{\prime} \times W_{i}^{\prime}}\}_{i=1}^{L_{f}}$, where $T$ is the length of the sequence, $H$ and $W$ are the height and width of the frames, $H_{i}^{\prime}$ and $W_{i}^{\prime}$ are those of the extracted feature maps, $D_{i}$ is the channel dimension, and $L_{f}$ is the number of extracted feature map scales. Following previous works~\cite{ehsanpour_eccv2020,li_iccv2021,tamura_eccv2022}, we employ the inflated 3D (I3D) network~\cite{carreira_cvpr2017} as a backbone network to embed local spatio-temporal contexts into feature maps. Note that we use only the RGB stream of I3D because group members are identified by their locations, which may not be predicted with the optical flow features. The feature map of each scale $\bm{z}_{i}^{\left(f\right)}$ is then mean-pooled over the temporal dimension and input to a projection convolution layer to suppress the channel dimension from $D_{i}$ to $D_{p}$. These two operations are done to reduce the computational costs of transformers in the subsequent process. In addition to the convolution layer for dimensionality reduction, a projection convolution layer that has a kernel size of $3 \times 3$ and stride of $2 \times 2$ is applied to the smallest feature map to add a feature map of a smaller scale.

The modified multi-scale feature maps are further reined with a deformable transformer encoder. Given a set of the modified multi-scale feature maps $\bm{Z}_{p} = \{\bm{z}_{i}^{\left(p\right)} \mid \bm{z}_{i}^{\left(p\right)} \in \mathbb{R}^{D_{p} \times H_{i}^{\prime} \times W_{i}^{\prime}}\}_{i=1}^{L_{f} + 1}$ from the projection convolution layers, a set of refined multi-scale feature maps $\bm{Z}_{e} = \{\bm{z}_{i}^{\left(e\right)} \mid \bm{z}_{i}^{\left(e\right)} \in \mathbb{R}^{D_{p} \times H_{i}^{\prime} \times W_{i}^{\prime}}\}_{i=1}^{L_{f} + 1}$ is obtained as $\bm{Z}_{e} = f_{enc}(\bm{Z}_{p}, \bm{P})$, where $f_{enc}(\cdot, \cdot)$ is stacked deformable transformer encoder layers and $\bm{P} = \{\bm{p}_{i} \mid \bm{p}_{i} \in \mathbb{R}^{D_{p} \times H_{i}^{\prime} \times W_{i}^{\prime}}\}_{i=1}^{L_{f} + 1}$ is a set of multi-scale positional encodings~\cite{zhu_iclr2021}, which is generated by adding learnable scale embeddings to two dimensional sinusoidal positional encodings. The multi-scale positional encodings supplement the attention modules in the encoder layers with position and scale information of each feature embedding to identify where it lies in the feature maps. The encoder enables feature embeddings to exchange information among those farther away from each other, and as a result, rich social group contexts can be acquired. The refined feature maps are shared by the individual and group recognition paths.

These enriched feature maps are fed into a deformable transformer decoder to aggregate social group features. Given a set of refined feature maps $\bm{Z}_{e}$ and learnable group queries $\bm{\mathcal{Q}}_{gr} = \{\bm{Q}_{i}^{(gr)} \mid \bm{Q}_{i}^{(gr)} = (\bm{q}_{i,j}^{(gr)} \mid \bm{q}_{i,j}^{(gr)} \in \mathbb{R}^{2D_{p}})_{j=1}^{N_{id}}\}_{i=1}^{N_{gr}}$, a set of social group features $\bm{\mathcal{H}}_{gr} = \{\bm{H}_{i}^{(gr)} \mid \bm{H}_{i}^{(gr)} = (\bm{h}_{i,j}^{(gr)} \mid \bm{h}_{i,j}^{(gr)} \in \mathbb{R}^{D_{p}})_{j=1}^{N_{id}}\}_{i=1}^{N_{gr}}$ is obtained as $\bm{\mathcal{H}}_{gr} = f_{dec}(\bm{Z_{e}}, \bm{\mathcal{Q}}_{gr})$, where $N_{gr}$ is the number of group queries, $N_{id}$ is the length of query embedding sequence in a group query, and $f_{dec}(\cdot, \cdot)$ is stacked deformable transformer decoder layers. The proposed method is designed so that each group query is assigned to at most one social group in a scene, and each query embedding captures at most one individual in the assigned social group. Generated feature embeddings are solely used to localize group members in a social group and combined with other embeddings in the same group to infer a group size and activity category. The total number of the embeddings $N_{gr}N_{id}$ becomes quite large because $N_{gr}$ must be large enough to capture social groups at any locations in frames, and $N_{id}$ must be larger than estimated maximum group size. This large number of embeddings complicates the query training and renders the decoder's self-attention ineffective. We describe an efficient way to create group queries with fewer learnable embeddings in Sec.~\ref{subsec:des_query} and a design to reduce the number of embeddings fed into the self-attention modules simultaneously in Sec.~\ref{subsec:des_attn}.

The generated feature embeddings are transformed into prediction results with FNNs in detection heads. Here we denote the localization results in normalized image coordinates and predicted group sizes in values normalized by $N_{id}$. Social group activity recognition is performed by predicting activity categories and identifying group members. The activity prediction is done as a typical multi-label classification problem with an activity prediction head. The identification is performed as group size estimation and member localization problems with a group size head and group member point head. The size head predicts the number of people in a target social group, and the point head indicates group members by localizing the centers of group members' bounding boxes. This design enables the proposed method to identify group members with simple point matching during inference as described in Sec.~\ref{subsec:mem_id}. The predictions of activity class probabilities $\{\hat{\bm{v}}_{i} \mid \hat{\bm{v}}_{i} \in [0, 1]^{N_{v}}\}_{i=1}^{N_{gr}}$, group sizes $\{\hat{s}_{i} \mid \hat{s}_{i} \in [0, 1]\}_{i=1}^{N_{gr}}$, and group member points $\{\hat{\bm{U}}_{i} \mid \hat{\bm{U}}_{i} = (\hat{\bm{u}}_{i,j} \mid \hat{\bm{u}}_{i,j} \in [0, 1]^{2})_{j = 1}^{N_{id}}\}_{i=1}^{N_{gr}}$ are obtained as $\hat{\bm{v}}_{i} = f_{v}(\bar{\bm{h}}_{i}^{(gr)})$, $\hat{s}_{i} = f_{s}(\bar{\bm{h}}_{i}^{(gr)})$, and $\hat{\bm{u}}_{i,j} = f_{u}(\bm{h}_{i,j}^{(gr)}, \bm{r}_{i,j}^{(gr)})$, where $N_{v}$ is the number of activity classes, $\bar{\bm{h}}_{i}^{(gr)} = \frac{1}{N_{id}}\sum_{j=1}^{N_{id}} \bm{h}_{i,j}^{(gr)}$, $f_{v}(\cdot)$, $f_{s}(\cdot)$, and $f_{u}(\cdot, \cdot)$ are FNNs with subsequent sigmoid functions in the detection heads for each prediction, and $\bm{r}_{i,j}^{(gr)} \in [0, 1]^2$ is a reference point, which is used in the same way as the localization in Deformable DETR~\cite{zhu_iclr2021}. The detailed implementations of the detection heads are described in the supplementary material of our previous paper~\cite{tamura_eccv2022}.

\subsection{Design of Group Queries}\label{subsec:des_query}

As mentioned in the previous section, the naive implementation of the group queries needs $N_{gr}N_{id}$ embeddings and thus increases a substantial number of learnable embedding compared to a transformer-based object detection framework~\cite{carion_eccv2020,zhu_iclr2021}. This increase may render training difficult because each query embedding is trained to have a target localization region as described in~\cite{carion_eccv2020}, though only a few of the embeddings are assigned to ground truth labels per sample during training. To solve this issue, we propose decomposing group queries into location queries $\bm{Q}_{lo} = \{\bm{q}_{i}^{(lo)} \mid \bm{q}_{i}^{(lo)} \in \mathbb{R}^{2D_{p}}\}_{i=1}^{N_{gr}}$ and shared layout queries $\bm{Q}_{la} = (\bm{q}_{i}^{(la)} \mid \bm{q}_{i}^{(la)} \in \mathbb{R}^{2D_{p}})_{i=1}^{N_{id}}$. Each query embedding in group queries is obtained by adding a location query embedding and layout query embedding as $\bm{q}_{i,j}^{(gr)} = \bm{q}_{i}^{(lo)} + \bm{q}_{j}^{(la)}$. The location queries provide representative locations of group queries, while the layout queries supply relative locations of individual queries in a group query. This design enables group queries to share the knowledge of a group member layout in a frame during training and thus alleviates the problem of the query embedding training. The comparison results of the naive and decomposed implementations on group activity recognition and social group activity recognition are described in Sec.~\ref{subsubsec:ana_qdes_gar} and Sec.~\ref{subsubsec:ana_qdes_sgar}, respectively.

\subsection{Design of Self-Attention Module}\label{subsec:des_attn}

\begin{figure*}[t]
    \centering
    \begin{subfigure}[t]{0.32\linewidth}
        \centering
        \raisebox{\height-6.3ex}{\includegraphics[width=0.89\linewidth]{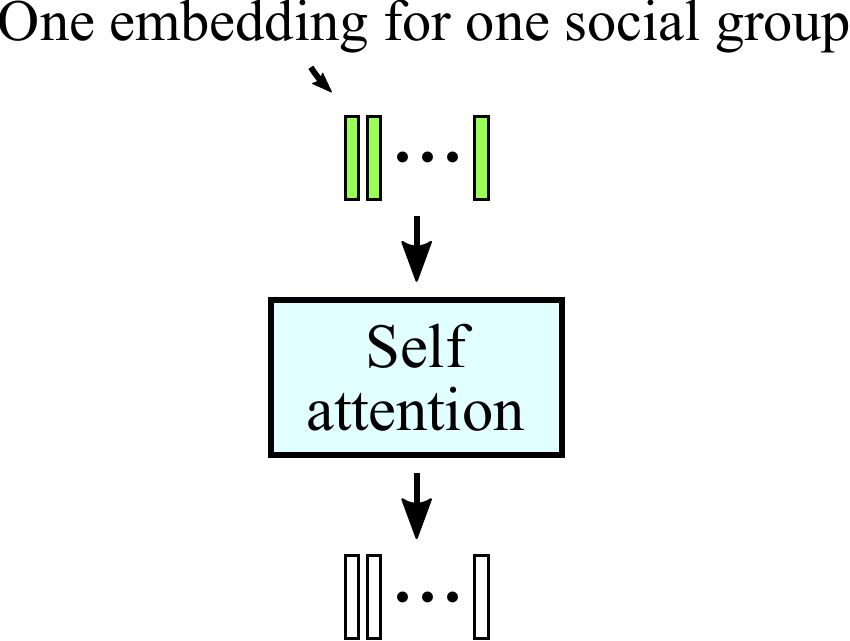}}
        \caption{Previous method~\cite{tamura_eccv2022}.}
        \label{fig:self_attn_prev}
    \end{subfigure}
    \hfill
    \begin{subfigure}[t]{0.32\linewidth}
        \centering
        \raisebox{\height-8.8ex}{\includegraphics[width=1.0\linewidth]{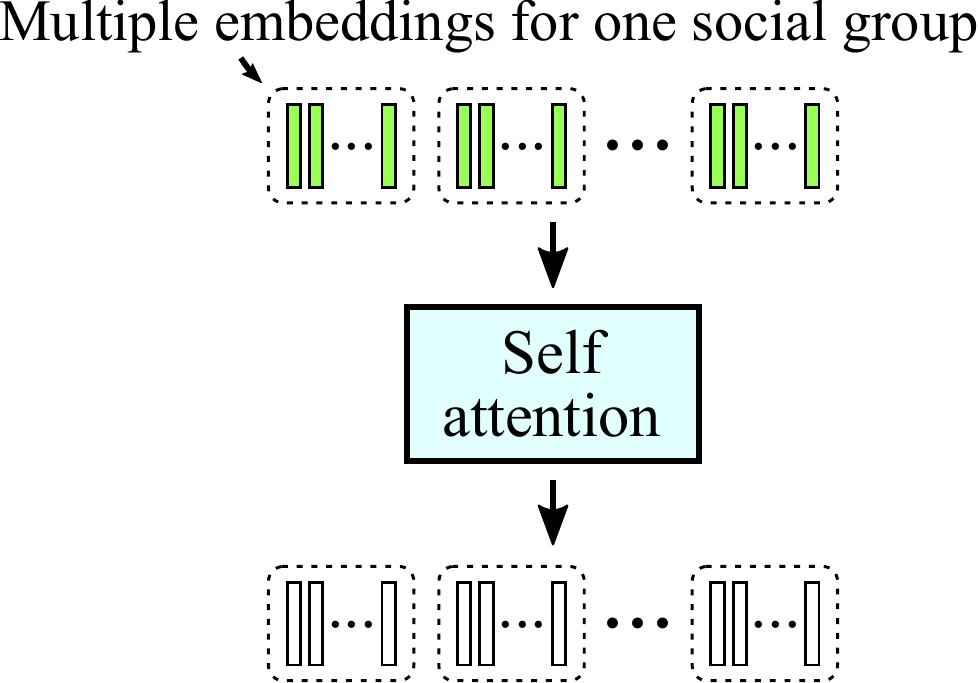}}
        \caption{Naive attention.}
        \label{fig:self_attn_naive}
    \end{subfigure}
    \hfill
    \begin{subfigure}[t]{0.32\linewidth}
        \centering
        \includegraphics[width=1.0\linewidth]{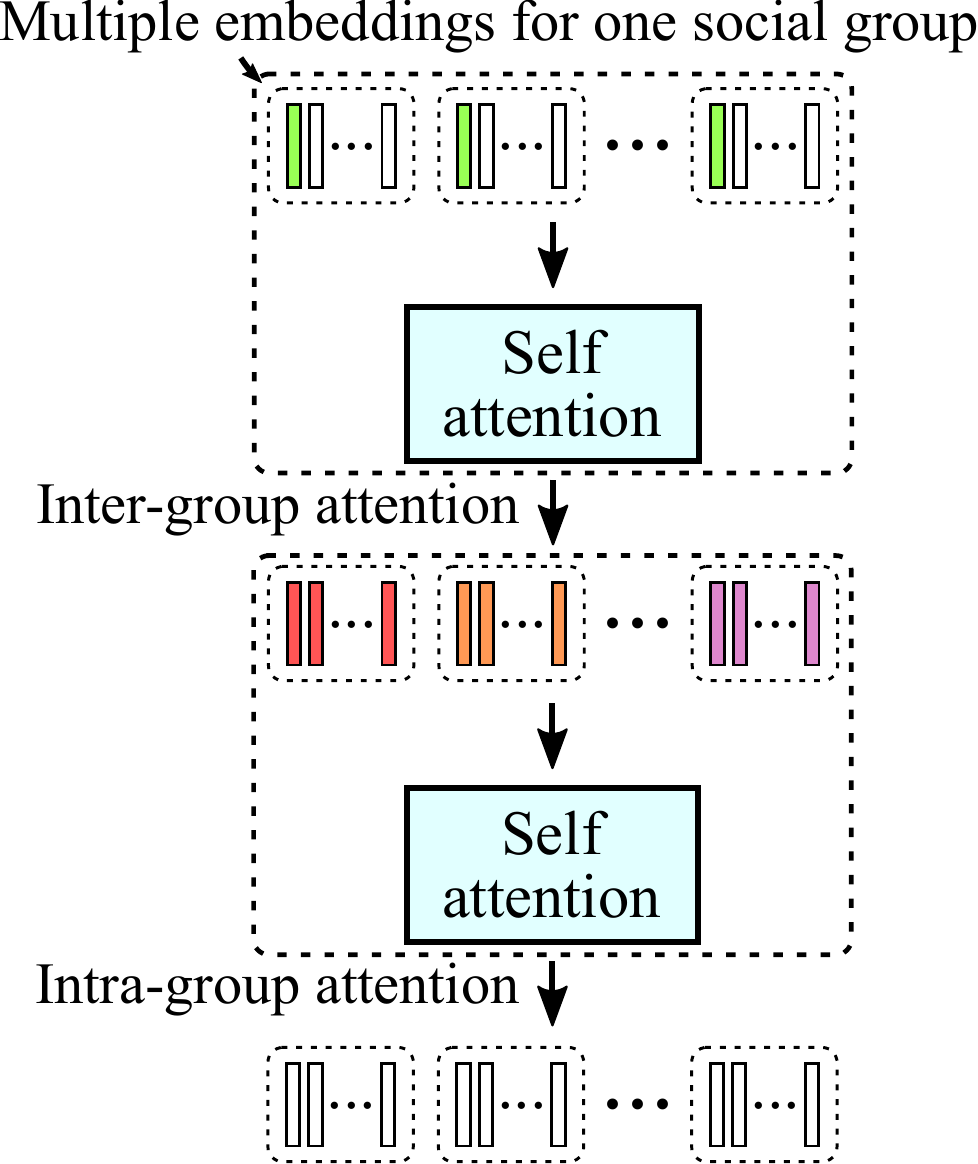}
        \caption{Divided attention.}
        \label{fig:self_attn_divide}
    \end{subfigure}
    \caption{Several design choices of the self-attention module in the transformer decoder. The colors of the embeddings indicate embedding sets. Only embeddings in the same set interact with each other using the attention mechanism. The embeddings of white colors are not fed into the attention module, which means that only one embedding in each group is fed into the first self-attention module in the divided attention implementation. Residual connections are omitted in this figure.}
    \label{fig:self_attn}
\end{figure*}

Since a significant amount of embeddings are fed into the self-attention modules in the transformer decoder, detailed communication of embeddings inside a group may not be performed with the naive implementation of the self-attention modules. However, such communication is essential to identify group members because detailed communication facilitates the non-duplicated assignment of the embeddings to group members and enables the embeddings to aggregate features of different group members for precise localization. To enable embeddings in the same group to communicate efficiently, we leverage the insight of divided attention~\cite{bertasius_icml2021} and propose splitting the self-attention into inter-group and intra-group self-attention. Figure~\ref{fig:self_attn} shows the naive and divided self-attention implementations in the transformer decoder. We also show the self-attention of our previous method~\cite{tamura_eccv2022} for the sake of easy comparison. Embedding sets are indicated by their colors except for the white color which indicates the embeddings that are not fed into the attention module. Embeddings interact with only those in the same sets by the attention mechanism. As shown in the figure, all the embeddings interact with each other in the naive implementation, whereas only the selected embeddings from each group interact with the first inter-group attention module, and then embeddings in the same groups interact with the second intra-group attention module in the divided attention implementation. The first attention module assigns group queries to social groups in frames with representative embeddings from each group. The first embeddings $\bm{q}_{i,0}^{(gr)}$ and $\bm{h}_{i,0}^{(gr)}$ in the embedding sequences are used for the representatives. The second attention module enables embeddings in the same group to decide the assignment between the embeddings and individuals in a group. This design enables embeddings to communicate closely inside a group and thus improve the performance of group member identification. The comparison results of the naive and divided attention implementations on group activity recognition and social group activity recognition are described in Sec.~\ref{subsubsec:ana_attn_gar} and Sec.~\ref{subsubsec:ana_attn_sgar}, respectively.

\subsection{Loss Calculation}\label{subsec:loss}

We view social group activity recognition as a set prediction problem and modify the training procedure of DETR~\cite{carion_eccv2020} for social group activity recognition. In this procedure, each social group in a ground truth set is assigned to a social group prediction in a prediction set with minimum cost matching between them. The costs are calculated with ground truth labels and predictions of activity class categories, group sizes, and group member locations. Given a ground truth set of social group activity recognition, the set is first padded with $\phi^{(gr)}$ (no activity) to change the set size to $N_{gr}$ for one-to-one matching between ground truth labels and predictions. With the padded ground truth set, the matching cost of the $i$-th element in the ground truth set and $j$-th element in the prediction set is calculated as follows:
\begin{align}
  \mathcal{H}^{(gr)}_{i, j} ={} & \mathbbm{1}_{\{i \not\in \bm{\Phi}^{(gr)}\}}\left[\eta_{v} \mathcal{H}^{(v)}_{i, j} + \eta_{s} \mathcal{H}^{(s)}_{i, j} + \eta_{u} \mathcal{H}^{(u)}_{i, j}\right], \\
  \mathcal{H}^{(v)}_{i, j} ={} & -\frac{\bm{v}^{T}_{i}\hat{\bm{v}}_{j} + \left(\bm{1} - \bm{v}_{i}\right)^{T}\left(\bm{1} - \hat{\bm{v}}_{j}\right)}{N_{v}}, \\
  \mathcal{H}^{(s)}_{i, j} ={} & \left\lvert s_{i} - \hat{s}_{j}\right\rvert, \\
  \mathcal{H}^{(u)}_{i, j} ={} & \frac{\sum_{k = 1}^{S_{i}} \left\|\bm{u}_{i,k} - \hat{\bm{u}}_{j,k}\right\|_{1}}{S_{i}},
\end{align}
where $\bm{\Phi}^{(gr)}$ is a set of indices that correspond to $\phi^{(gr)}$, $\bm{v}_{i} \in \{0, 1\}^{N_{v}}$ is a ground truth activity label, $s_{i} \in [0, 1]$ is a ground truth group size normalized with $N_{id}$, $S_{i}$ is an unnormalized ground truth group size, $\bm{u}_{i,k} \in [0, 1]^{2}$ is a ground-truth group member point normalized with the image size, and $\eta_{\{v, s, u\}}$ are hyper-parameters for balancing the costs. Group member points in the sequence $\bm{U}_{i} = (\bm{u}_{i,k})^{S_{i}}_{k = 1}$ are sorted in ascending order along $X$ coordinates as seen from the image of the group recognition result in Fig.~\ref{fig:overview}. We use this arrangement because group members are typically seen side by side at the same vertical positions in an image, and the order of group member points is clear from their positions, which simplifies the prediction. We verified the effectiveness of this arrangement in our previous work~\cite{tamura_eccv2022}. The Hungarian algorithm~\cite{kuhn_naval1955} is applied to the calculated costs to find the optimal assignment as $\hat{\omega}^{(gr)} = \argmin_{\omega \in \bm{\Omega}_{N_{gr}}}{\sum_{i=1}^{N_{gr}}{\mathcal{H}^{(gr)}_{i,\omega(i)}}}$, where $\bm{\Omega}_{N_{gr}}$ is the set of all possible permutations of $N_{gr}$ elements.

After the minimum cost matching, the training loss $\mathcal{L}_{gr}$ is calculated between matched ground truth labels and predictions as follows:
\begin{align}
  \mathcal{L}_{gr} ={} & \lambda_{v}\mathcal{L}_{v} + \lambda_{s}\mathcal{L}_{s} + \lambda_{u}\mathcal{L}_{u}, \\
  \mathcal{L}_{v} ={} & \frac{1}{\lvert\bar{\bm{\Phi}}^{(gr)}\rvert} \sum_{i=1}^{N_{gr}}\left[
          \mathbbm{1}_{\{i \not\in \bm{\Phi}^{(gr)}\}}l_{f}\left(\bm{v}_{i}, \hat{\bm{v}}_{\hat{\omega}^{(gr)}\left(i\right)}\right) + \right. \nonumber \\
          &\quad\quad\quad\quad \left. \mathbbm{1}_{\{i \in \bm{\Phi}^{(gr)}\}}l_{f}\left(\bm{0}, \hat{\bm{v}}_{\hat{\omega}^{(gr)}\left(i\right)}\right)\right], \\
  \mathcal{L}_{s} ={} & \frac{1}{\lvert\bar{\bm{\Phi}}^{(gr)}\rvert} \sum_{i=1}^{N_{gr}} \mathbbm{1}_{\{i \not\in \bm{\Phi}^{(gr)}\}}\left\lvert s_i - \hat{s}_{\hat{\omega}^{(gr)}\left(i\right)}\right\rvert, \\
  \mathcal{L}_{u} ={} & \frac{1}{\lvert\bar{\bm{\Phi}}^{(gr)}\rvert} \sum_{i=1}^{N_{gr}} \sum_{j=1}^{S_{i}} \mathbbm{1}_{\{i \not\in \bm{\Phi}^{(gr)}\}} \left\|\bm{u}_{i,j} - \hat{\bm{u}}_{\hat{\omega}^{(gr)}\left(i\right),j}\right\|_{1},
\end{align}
where $\lambda_{\{v, s, u\}}$ are hyper-parameters for balancing the losses and $l_{f}(\cdot, \cdot)$ is the element-wise focal loss function~\cite{lin_iccv2017} whose hyper-parameters are set as described in~\cite{zhou_arxiv2019}.

\subsection{Group Member Identification}\label{subsec:mem_id}

During inference, bounding boxes of group members must be obtained to identify them. We acquire the bounding boxes by transforming the predicted group sizes and group member points into indices of the elements in the individual prediction set. This transformation is performed by minimum cost matching between the predicted group member points from the group detection heads and bounding box centers from the individual detection heads. The matching cost between $i$-th group member point of $k$-th social group prediction and $j$-th individual prediction is calculated as follows:
\begin{equation}
  \mathcal{H}^{(gm, k)}_{i, j} = \frac{\left\|\hat{\bm{u}}_{k,i} - f_{cent}\left(\hat{\bm{b}}_{j}\right)\right\|_{2}}{\hat{c}_{j}},
\end{equation}
where $\hat{\bm{b}}_{j} \in [0, 1]^{4}$ is a predicted bounding box of an individual, $\hat{c}_{j} \in [0, 1]$ is a detection score of the individual, and $f_{cent}(\cdot)$ is a function that calculates a bounding box center. The Hungarian algorithm~\cite{kuhn_naval1955} is used to find the optimal assignment between group member points and bounding box centers as $\hat{\omega}^{(gm, k)} = \argmin_{\omega \in \bm{\Omega}_{N_{b}}}{\sum_{i=1}^{\left\lfloor \hat{s}_{k} N_{id}\right\rceil}{\mathcal{H}^{(gm, k)}_{i,\omega(i)}}}$, where $N_{b}$ is the number of elements in the individual recognition result set, $\bm{\Omega}_{N_{b}}$ is the set of all possible permutations of $N_{b}$ elements, and $\lfloor \cdot\rceil$ rounds an input value to the nearest integer. Finally, the index set of individuals for $k$-th social group prediction is obtained as $\bm{G}_{k} = \{\hat{\omega}^{(gm, k)}\left(i\right)\}^{\left\lfloor \hat{s}_{k} N_{id}\right\rceil}_{i=1}$. The minimum cost matching prevents more than one group member point from being assigned to the same individuals, achieving slightly better performance than calculating the closest bounding box center for each group member point.

\section{Experiments}\label{sec:exp}

\subsection{Datasets}

Following existing works~\cite{ehsanpour_eccv2020,li_iccv2021}, we use two publicly available benchmark datasets to show the effectiveness of the proposed method.
The first dataset is the Volleyball dataset. The Volleyball dataset contains 4,830 videos of 55 volleyball matches, which are split into 3,493 training videos and 1,337 test videos. The only center frame of each video is manually annotated with bounding boxes, actions of 9 categories, and one group activity of 8 categories. The bounding boxes of other frames are obtained with a visual tracker. In addition to the original annotations, we use an extra annotation set provided by Sendo and Ukita~\cite{sendo_mva2019} because the original annotations do not contain group member information. We combine the original annotations with the group annotations in the extra set for training our model. Note that annotations other than the group annotations in the extra set are not used for a fair comparison.
The second dataset is the Collective Activity dataset. The Collective Activity dataset contains 44 life scene videos, which are split into 32 training videos and 12 test videos. The videos are manually annotated every ten frames with bounding boxes and actions of 6 categories. Group activity is defined as the most frequent action in a frame. We use Ehsanpour~\etal's annotations~\cite{ehsanpour_eccv2020} in addition to the original annotations for group information.
Note that the Volleyball dataset is used for analyses in our experiments because the Collective Activity dataset has limited diversity in its data and thus is unsuitable for obtaining reliable results.

\subsection{Evaluation Metrics}

Our experiments are broadly divided into those of group activity recognition and social group activity recognition.
The evaluation of group activity recognition is conducted with classification accuracy as an evaluation metric, where one predicted group activity is confirmed if it matches a ground truth label for each scene. Since the proposed method is designed to output predictions of multiple social groups in a scene, we select one activity with the highest probability from all the predicted activities and compare it to the ground truth label.
The evaluation of social group activity recognition is performed with different metrics for the Volleyball and Collective Activity datasets because each scene in the Volleyball dataset is ensured to have only one social group, while that in the Collective Activity dataset contains one or more social groups.
    The metric of social group activity recognition accuracy is used in the evaluation of the Volleyball dataset, where a prediction is considered to be correct if the predicted activity matches the ground truth activity and the predicted bounding boxes have intersection over unions (IoUs) larger than 0.5 with the corresponding ground truth boxes of the group members. We select one social group that has an activity prediction with the highest probability in each scene and compare it to the ground truth social group.
    In the evaluation of the Collective Activity dataset, mAP is used as an evaluation metric. Predictions are judged as true positives if the predicted activities are correct and all the predicted boxes have IoUs larger than 0.5 with the corresponding ground truth boxes of the group members.

\subsection{Implementation Details}

We use the RGB stream of I3D~\cite{carreira_cvpr2017} as a backbone network, whose feature maps of \textit{Mixed\_3c}, \textit{Mixed\_4f}, and \textit{Mixed\_5c} layers are fed into the deformable transformers. The hyper-parameters of the deformable transformers are set in accordance with those of Deformable DETR~\cite{zhu_iclr2021}, where $L_{f} = 3$ and $D_{p} = 256$. The I3D network is initialized with the parameters trained on the Kinetics dataset~\cite{kay_arxiv2017}, and the deformable transformers are initialized with the parameters trained on the COCO dataset~\cite{lin_eccv2014}. The number of group queries $N_{gr}$ and the number of query embeddings in a group query $N_{id}$ are set to 300 and 12, respectively. The length of the sequence $T$ is set to 9. For the training, we use the AdamW~\cite{loshchiloy_iclr2019} optimizer with a batch size of 16, whose initial learning rate and weight decay are both set to $10^{-4}$. Our model is trained for 120 epochs with the learning rate decay at 100 epochs. The hyper-parameters for balancing the costs and losses are set as $\eta_{v} = \lambda_{v} = 2$, $\eta_{s} = \lambda_{s} = 1$, and $\eta_{u} = \lambda_{u} = 5$. To augment the training data, we randomly shift frames in the temporal direction and use bounding boxes from visual trackers as ground truth bounding boxes when a non-annotated frame is at the center. We also augment the training data by random horizontal flipping, scaling, and cropping. Auxiliary losses are used to boost the performance following the DETR's training~\cite{carion_eccv2020}.

Some hyper-parameters are set to different values than those written above in the evaluation of the Collective Activity dataset. In the evaluation of group activity recognition, training epochs are set to 10, and the learning rate is decayed after 5 epochs because the losses converge in a few epochs due to the limited diversity of the scenes in the dataset. In the evaluation of social group activity recognition, the length of the sequence $T$ is set to 17 following the setting of Ehsanpour~\etal's work~\cite{ehsanpour_eccv2020}.

\subsection{Group Activity Recognition}

\subsubsection{Comparison against State-of-The-Art}\label{subsubsec:comp_sota_gar}

\begin{table*}[t]
    \caption{Comparison against state-of-the-art methods on group activity recognition. The values with and without the brackets demonstrate the performances in the ground-truth-based and detection-based settings, respectively. The performances of individual action recognition are shown for future reference.}
    \label{table:comp_groupact}
    \centering
    \setlength{\tabcolsep}{8.5pt}
    \begin{threeparttable}
        \begin{tabular}{@{}lcccccccccc@{}}
            \toprule
            &&& \multicolumn{4}{c}{Volleyball} & \multicolumn{4}{c}{Collective Activity} \\
            \cmidrule(lr){4-7}\cmidrule(lr){8-11}
            Method & Flow & Pose & \multicolumn{2}{c}{Activity} & \multicolumn{2}{c}{Action} & \multicolumn{2}{c}{Activity} & \multicolumn{2}{c}{Action} \\
            \midrule
            SIM~\cite{deng_cvpr2016} &&& -- & (\hspace{0.6em}--\hspace{0.6em}) & -- & (\hspace{0.6em}--\hspace{0.6em}) & -- & (81.2) & -- & (\hspace{0.6em}--\hspace{0.6em}) \\
            HDTM~\cite{ibrahim_cvpr2016} &&& -- & (81.9) & -- & (\hspace{0.6em}--\hspace{0.6em}) & -- & (81.5) & -- & (\hspace{0.6em}--\hspace{0.6em}) \\
            CERN~\cite{shu_cvpr2017} &&& -- & (83.3) & -- & (69.1) & -- & (87.2) & -- & (\hspace{0.6em}--\hspace{0.6em}) \\
            SSU~\cite{bagautdinov_cvpr2017} &&& 86.2 & (90.6) & -- & (81.8) & -- & (\hspace{0.6em}--\hspace{0.6em}) & -- & (\hspace{0.6em}--\hspace{0.6em}) \\
            SBGAR~\cite{li_iccv2017} & \checkmark && -- & (66.9) & -- & (\hspace{0.6em}--\hspace{0.6em}) & -- & (86.1) & -- & (\hspace{0.6em}--\hspace{0.6em}) \\
            HACN~\cite{kong_icassp2018} &&& -- & (85.1) & -- & (\hspace{0.6em}--\hspace{0.6em}) & -- & (84.3) & -- & (\hspace{0.6em}--\hspace{0.6em}) \\
            HRN~\cite{ibrahim_eccv2018} &&& -- & (89.5) & -- & (\hspace{0.6em}--\hspace{0.6em}) & -- & (\hspace{0.6em}--\hspace{0.6em}) & -- & (\hspace{0.6em}--\hspace{0.6em}) \\
            stagNet~\cite{qi_eccv2018} &&& 87.6 & (89.3) & -- & (\hspace{0.6em}--\hspace{0.6em}) & 87.9 & (89.1) & -- & (\hspace{0.6em}--\hspace{0.6em}) \\
            ARG~\cite{wu_cvpr2019} &&& 91.5 & (92.5) & 39.8 & (83.0) & 86.1 & (88.1) & 49.6 & (77.3) \\
            CRM~\cite{azar_cvpr2019} & \checkmark && -- & (93.0) & -- & (\hspace{0.6em}--\hspace{0.6em}) & -- & (85.8) & -- & (\hspace{0.6em}--\hspace{0.6em}) \\
            PRL~\cite{hu_cvpr2020} &&& -- & (91.4) & -- & (\hspace{0.6em}--\hspace{0.6em}) & -- & (\hspace{0.6em}--\hspace{0.6em}) & -- & (\hspace{0.6em}--\hspace{0.6em}) \\
            Actor-Transformers~\cite{gavrilyuk_cvpr2020} & \checkmark& \checkmark & -- & (94.4) & -- & (85.9) & -- & (92.8) & -- & (\hspace{0.6em}--\hspace{0.6em}) \\
            Ehsanpour~\etal~\cite{ehsanpour_eccv2020} & \checkmark && 93.0 & (93.1) & 41.8 & (83.3) & 89.4 & (89.4) & 55.9 & (78.3) \\
            Pramono~\etal~\cite{pramono_eccv2020} & \checkmark & \checkmark & -- & (95.0) & -- & (83.1) & -- & (95.2) & -- & (\hspace{0.6em}--\hspace{0.6em}) \\
            DIN~\cite{yuan_iccv2021} &&& -- & (93.6) & -- & (\hspace{0.6em}--\hspace{0.6em}) & -- & (95.9) & -- & (\hspace{0.6em}--\hspace{0.6em}) \\
            GroupFormer~\cite{li_iccv2021} & \checkmark & \checkmark & 95.0\tnote{$\ast$} & (95.7) & -- & (85.6) & 85.2\tnote{$\ast$} & (87.5\tnote{$\star$} /96.3) & -- & (\hspace{0.6em}--\hspace{0.6em}) \\
            \multicolumn{11}{@{}c@{}}{\makebox[\linewidth]{\dashrule[black]}} \\
            Ours prev.~\cite{tamura_eccv2022} &&& \textbf{96.0} & (\hspace{0.6em}--\hspace{0.6em}) & 65.0 & (\hspace{0.6em}--\hspace{0.6em}) & \textbf{96.5} & (\hspace{0.6em}--\hspace{0.6em}) & 64.9 & (\hspace{0.6em}--\hspace{0.6em}) \\
            Ours &&& 95.4 & (\hspace{0.6em}--\hspace{0.6em}) & \textbf{65.5} & (\hspace{0.6em}--\hspace{0.6em}) & 96.3 & (\hspace{0.6em}--\hspace{0.6em}) & \textbf{66.6} & (\hspace{0.6em}--\hspace{0.6em}) \\
            \bottomrule
        \end{tabular}
        \begin{tablenotes}\footnotesize
            \item[$\ast$] We evaluated the performance with the publicly available source codes.
            \item[$\star$] We evaluated but were not able to reproduce the reported accuracy because the configuration file for the Collective Activity dataset is not publicly available.
        \end{tablenotes}
    \end{threeparttable}
\end{table*}

To demonstrate the effectiveness of the proposed method, we first compare the method with state-of-the-art group activity recognition methods. Table~\ref{table:comp_groupact} shows the comparison results. We also show the performances of individual action recognition for future reference. The values with the brackets show the performances of the methods whose region features are extracted with ground truth bounding boxes, while those without brackets illustrate the performances with predicted boxes. Since we do not use any ground truth bounding boxes for the prediction, we list the performance of our methods under those of the detection-based setting. Several detection-based performances are not reported in the table because existing works typically use ground-truth boxes for the evaluation. To fairly compare the proposed method with a state-of-the-art method, we evaluate GroupFormer~\cite{li_iccv2021}, which is the strongest baseline of group activity recognition, with the predicted boxes of Deformable DETR~\cite{zhu_iclr2021}. Note that Deformable DETR is fine-tuned on each dataset for a fair comparison, which demonstrates 90.8 and 90.2 mAP on the Volleyball and Collective Activity datasets, respectively.

As seen from the table, the proposed method achieves state-of-the-art performance on both Volleyball and Collective Activity datasets in the detection-based setting. The performance is even competitive when compared to those of the existing methods in the ground-truth-based setting. It is worth mentioning that recent state-of-the-art methods leverage either or both flow and pose features to enhance the performances, while the proposed method uses only features extracted from RGB images. These results indicate that our feature extraction is more effective than extracting region features.

As seen from the comparison with the previous work, using multiple embeddings for recognizing one social group slightly degrades the performance. Since group activity recognition does not require localizing group members, the inter-group communication of embeddings is probably more critical than the intra-group communication. The fewer embeddings and single self-attention module simplify the inter-group communication and thus achieve better performance on group activity recognition. This assumption is verified in Sec.~\ref{subsubsec:ana_attn_gar}.

\subsubsection{Analysis of Query Design}\label{subsubsec:ana_qdes_gar}

\begin{table}[t]
    \caption{Comparison of the naive and decomposed group query implementation on group activity recognition.}
    \label{table:comp_groupact_divquery}
    \centering
    \begin{tabular}{@{}lc@{}}
        \toprule
        Method & Accuracy \\
        \midrule
        Naive query implementation & 95.1 \\
        Decomposed query implementation & 95.4 \\
        \bottomrule
    \end{tabular}
\end{table}

To analyze the effectiveness of the decomposed group query implementation described in Sec.~\ref{subsec:des_query}, we compare the performances of the naive and decomposed query implementation on group activity recognition. Table~\ref{table:comp_groupact_divquery} shows the comparison results. As depicted in the table, the decomposed implementation slightly improves the performance. Although group activity recognition does not require localizing group members, sharing a layout between group queries helps identify feature locations and thus has a benefit for generating effective features. Accordingly, the decomposed implementation demonstrates better performance than the naive implementation.

\subsubsection{Analysis of-Self Attention Design}\label{subsubsec:ana_attn_gar}

\begin{table}[t]
    \caption{Comparison of three divided self-attention designs on group activity recognition. Inter-group and intra-group refer to the attention modules depicted in Fig.~\ref{fig:self_attn_divide}.}
    \label{table:comp_groupact_divattndes}
    \centering
    \begin{tabular}{@{}lc@{}}
        \toprule
        Method & Accuracy \\
        \midrule
        Only inter-group & 94.1 \\
        Intra-group $\rightarrow$ inter-group & 94.5 \\
        Inter-group $\rightarrow$ intra-group & 95.4 \\
        \bottomrule
    \end{tabular}
\end{table}

Before analyzing the effectiveness of the divided self-attention, we test several designs of the divided self-attention module to figure out the optimal design of it. We compare the one introduced in Sec.~\ref{subsec:des_attn} with two variants, one of which has only the inter-group attention module and the other of which changes the order of the inter-group and intra-group attention modules. Table~\ref{table:comp_groupact_divattndes} shows the comparison results. As shown in the table, the design in Fig.~\ref{fig:self_attn_divide} shows better performance than the other two variants. The lowest performance of the variant without the intra-group attention is quite reasonable because the variant disables communication between embeddings of the same group and thus renders the cooperative feature extraction difficult. In terms of the order of the attention modules, applying inter-group attention and then intra-group attention is more sensible because the information of the assignment between group queries and social groups can be distributed to each embedding by that order. This information distribution is probably the reason why the design in Fig.~\ref{fig:self_attn_divide} shows better performance than the variant having the opposite order. We observe the same trend of the performances in the evaluation of social group activity recognition, which is described in Sec.~\ref{subsubsec:ana_attn_sgar}, and thus conclude that the design in Fig.~\ref{fig:self_attn_divide} is the optimal one. We use the design for the divided attention in our experiments unless otherwise noted.

\begin{table}[t]
    \caption{Comparison of self-attention designs on group activity recognition.}
    \label{table:comp_groupact_divattn}
    \centering
    \setlength{\tabcolsep}{12.2pt}
    \begin{tabular}{@{}lccc@{}}
        \toprule
        Method & $N_{id}$ & Divided attn. & Accuracy \\
        \midrule
        Prev.~\cite{tamura_eccv2022} & 1 & & 96.0 \\
        \multicolumn{4}{@{}c@{}}{\makebox[\linewidth]{\dashrule[black]}} \\
        Divided attn. & 12 & \checkmark & 95.4 \\
        Naive attn. & 6 & & 95.7 \\
        \bottomrule
    \end{tabular}
\end{table}

As described in Sec.~\ref{subsubsec:comp_sota_gar}, the proposed method shows slightly worse performance than the previous method~\cite{tamura_eccv2022} on group activity recognition probably because embeddings in the previous method can have better inter-group communicate due to the fewer embeddings and single attention module. To verify this assumption, we compare the divided self-attention implementation to the naive self-attention implementation with fewer embeddings for one social group. Table~\ref{table:comp_groupact_divattn} shows the comparison results. The naive implementation with fewer embeddings outperforms the divided attention implementation. Since the naive implementation calculates attention values in a single attention module, communication between group queries is simplified if the number of embeddings is limited, resulting in better performance than the divided attention implementation. The results suggest that simple implementations of the self-attention modules in the transformer decoder are suitable for group activity recognition. However, both the previous method and the naive implementation have difficulty in identifying group members and thus have lower performance than the divided attention implementation on social group activity recognition. This observation is confirmed in Sec.~\ref{subsubsec:ana_attn_sgar}. 

\subsection{Social Group Activity Recognition}

\subsubsection{Comparison against State-of-The-Art}\label{subsubsec:comp_sgar}

To show the effectiveness of the proposed method on social group activity recognition, we compare the proposed method with Ehsanpour~\etal's method~\cite{ehsanpour_eccv2020}, which is a state-of-the-art method of social group activity recognition, GroupFormer~\cite{li_iccv2021}, which is the strongest baseline of group activity recognition, and our previous method~\cite{tamura_eccv2022}. We implement Ehsanpour~\etal's method based on our best understanding and evaluate the performance of the Volleyball dataset because they do not provide the source codes and do not report the results of the dataset. Since GroupFormer does not contain a scheme to identify group members, we train Deformable DETR~\cite{zhu_iclr2021} with only bounding boxes of group members to detect people involved in activities, which shows the detection performance of 87.1 mAP. This approach can recognize at most one social group in a scene and thus is tested only on the Volleyball dataset.

\begin{table*}[t]
    \caption{Comparison against state-of-the-art methods on social group activity recognition with the Volleyball dataset.}
    \label{table:comp_socialgroup_volley}
    \centering
    \setlength{\tabcolsep}{6.5pt}
    \begin{threeparttable}
        \begin{tabular}{@{}lccccccccccc@{}}
            \toprule
            & & & & \multicolumn{4}{c}{Right} & \multicolumn{4}{c}{Left} \\
            \cmidrule(lr){5-8}\cmidrule(lr){9-12}
            Method & Flow & Pose & Accuracy & Set & Spike & Pass & Winpoint & Set & Spike & Pass & Winpoint \\
            \midrule
            Ehsanpour~\etal~\cite{ehsanpour_eccv2020}\tnote{\dag} & \checkmark & & 44.5 & 17.2 & \textbf{74.0} & 49.0 & 29.9 & 19.7 & \textbf{79.6} & 25.0 & 28.4 \\
            GroupFormer~\cite{li_iccv2021}\tnote{\ddag} & \checkmark & \checkmark & 48.8 & 25.0 & 56.6 & 59.0 & 51.7 & 31.5 & 55.3 & 58.8 & 51.0 \\
            \multicolumn{12}{@{}c@{}}{\makebox[\linewidth]{\dashrule[black]}} \\
            Ours prev.~\cite{tamura_eccv2022} & & & 60.6 & 35.9 & 68.2 & \textbf{81.9} & 50.6 & 50.6 & 53.6 & 74.3 & 56.9 \\
            Ours & & & \textbf{64.4} & \textbf{40.6} & 66.5 & 80.5 & \textbf{60.9} & \textbf{55.4} & 61.5 & \textbf{77.0} & \textbf{67.6} \\
            \bottomrule
        \end{tabular}
        \begin{tablenotes}\footnotesize
            \item[\dag] Because the source codes are not publicly available, we implemented their algorithm based on our best understanding and evaluated the performance.
            \item[\ddag] We trained a group member detector and evaluated the performance with publicly available source codes.
        \end{tablenotes}
    \end{threeparttable}
\end{table*}

The comparison results on the Volleyball dataset are listed in Table~\ref{table:comp_socialgroup_volley}. The proposed method outperforms the existing two methods and our previous method. In particular, the proposed method increases the accuracy of the activity ``Winpoint", which typically involves more people than the other activities, by over 10 points compared to the previous method. These results indicate the advantage of using multiple embeddings for recognizing one social group. Since the previous method encodes the location information of all the group members into one embedding, group member identification becomes difficult if group sizes are large. Meanwhile, the proposed method assigns one embedding to each group member and thus does not have such a problem. This embedding assignment is the reason why the proposed method demonstrates better performance than the previous method on social group activity recognition, even though the previous one shows better performance on group activity recognition. This observation is further analyzed in Sec.~\ref{subsubsec:ana_gsizeloc}.

\begin{table*}[t]
 \caption{Comparison against state-of-the-art methods on social group activity recognition with the Collective Activity dataset.}
 \label{table:comp_socialgroup_collective}
 \centering
 \setlength{\tabcolsep}{11pt}
 \begin{tabular}{@{}lcccccccc@{}}
  \toprule
  Method & Flow & Pose & mAP & Crossing & Waiting & Queueing & Walking & Talking \\
  \midrule
  Ehsanpour~\etal~\cite{ehsanpour_eccv2020} & \checkmark & & \textbf{51.3} & -- & -- & -- & -- & -- \\
  \multicolumn{9}{@{}c@{}}{\makebox[\linewidth]{\dashrule[black]}} \\
  Ours prev.~\cite{tamura_eccv2022} & & & 46.0 & 49.2 & 64.5 & 54.1 & 55.6 & 6.56 \\
  Ours & & & 47.0 & 42.3 & 69.1 & 68.0 & 53.1 & 2.25 \\
  \bottomrule
 \end{tabular}
\end{table*}

Table~\ref{table:comp_socialgroup_collective} shows the comparison results on the Collective Activity dataset. The proposed method demonstrates slightly better performance than the previous method. However, the proposed method still cannot achieve Ehsanpour~\etal's performance. As analyzed in the previous work~\cite{tamura_eccv2022}, the low performances of our methods are attributed to the performances of the activity ``talking". \qty{86}{\percent} of samples with the activity ``talking" have the group sizes of four in the test set, while the training data has only 57 samples whose group sizes are four, which is \qty{0.8}{\percent} of the training data. Since our group member identification is based on group-size learning, the number of training samples for each group size affects the performance. Ehsanpour~\etal's method, on the other hand, uses clustering of individual embeddings to split individuals into social groups. This approach is not affected by the number of training samples for each group size and rather relies on the quality of the embeddings to identify group members. The Collective Activity dataset is suitable for this approach because individual actions, which are likely to impact individual embeddings, are always aligned with group activities due to the definition of the group activities in the dataset. However, embedding clustering may not work well if activities do not have specific sets of individual actions. This assumption is analyzed with the Volleyball dataset in Sec.~\ref{subsubsec:ana_indrec}.

\subsubsection{Analysis of Groups Sizes and Member Locations}\label{subsubsec:ana_gsizeloc}

\begin{figure*}[!htb]
    \centering
    \begin{subfigure}[b]{0.44\linewidth}
        \centering
        \includegraphics[keepaspectratio,width=\textwidth]{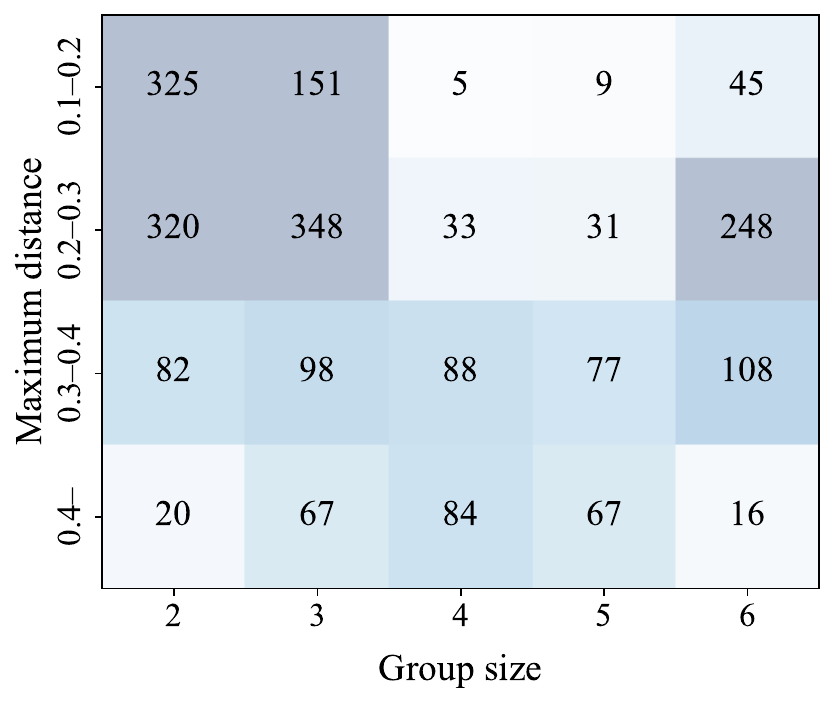}
        \caption{\#Training samples}
        \label{fig:dist_size_train}
    \end{subfigure}
    \hspace{5ex}
    \begin{subfigure}[b]{0.44\linewidth}
        \centering
        \includegraphics[keepaspectratio,width=\textwidth]{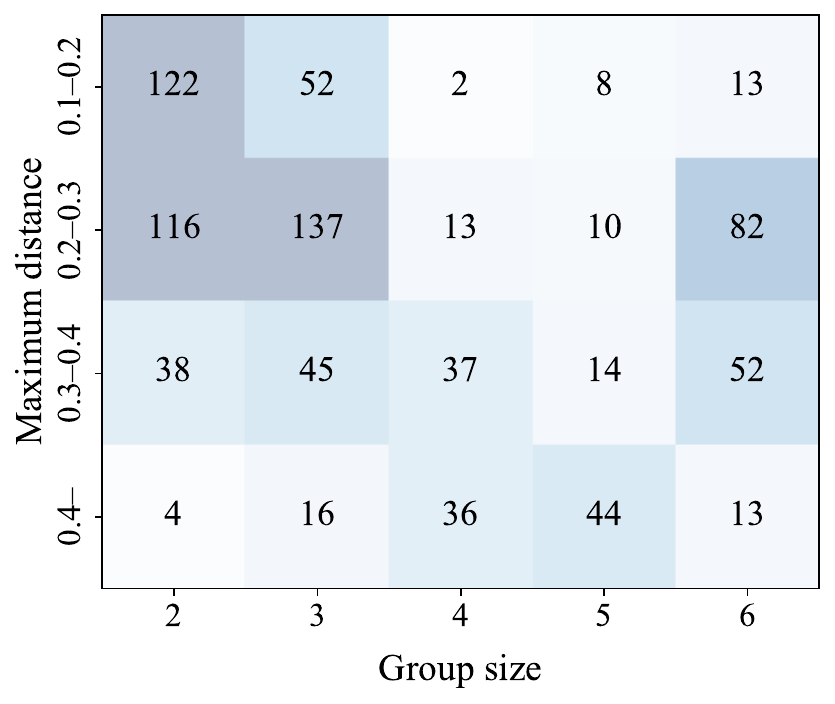}
        \caption{\#Test samples}
        \label{fig:dist_size_test}
    \end{subfigure} \\
    \begin{subfigure}[b]{0.44\linewidth}
        \centering
        \includegraphics[keepaspectratio,width=\textwidth]{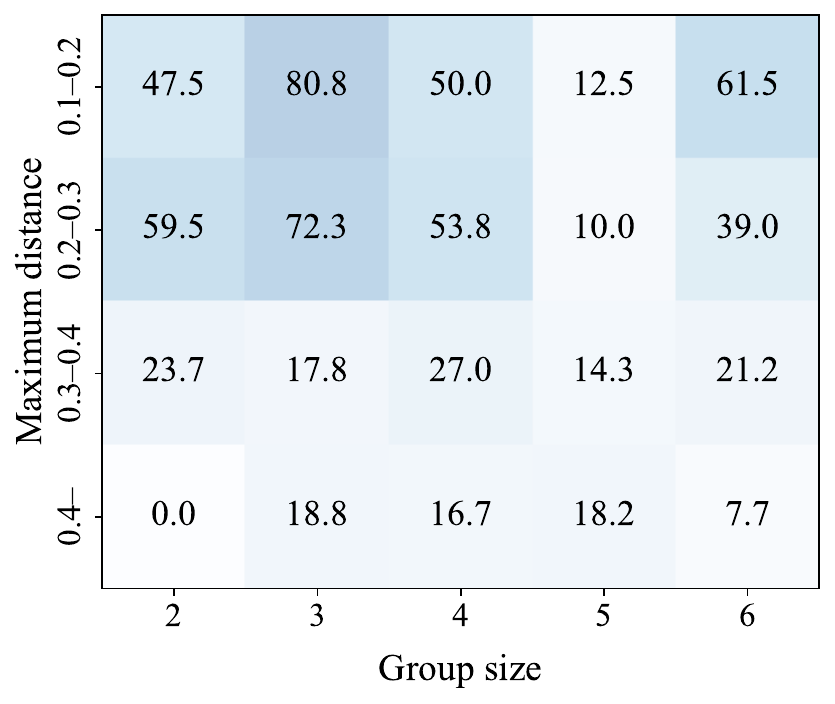}
        \caption{Ehsanpour~\etal~\cite{ehsanpour_eccv2020}}
        \label{fig:dist_size_ehsam}
    \end{subfigure}
    \hspace{5ex}
    \begin{subfigure}[b]{0.44\linewidth}
        \centering
        \includegraphics[keepaspectratio,width=\textwidth]{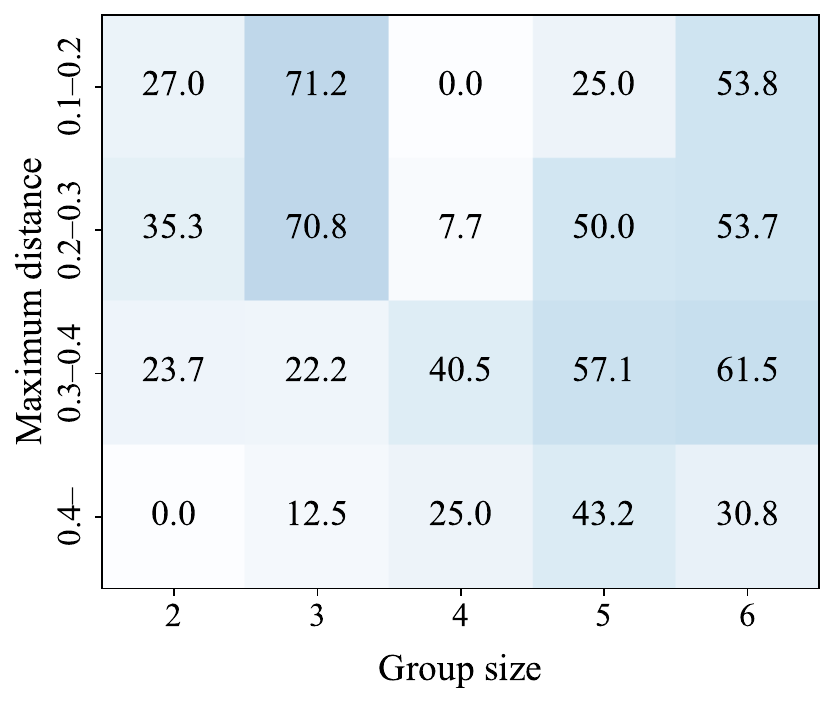}
        \caption{GroupFormer~\cite{li_iccv2021}}
        \label{fig:dist_size_gf}
    \end{subfigure} \\
    \begin{subfigure}[b]{0.44\linewidth}
        \centering
        \includegraphics[keepaspectratio,width=\textwidth]{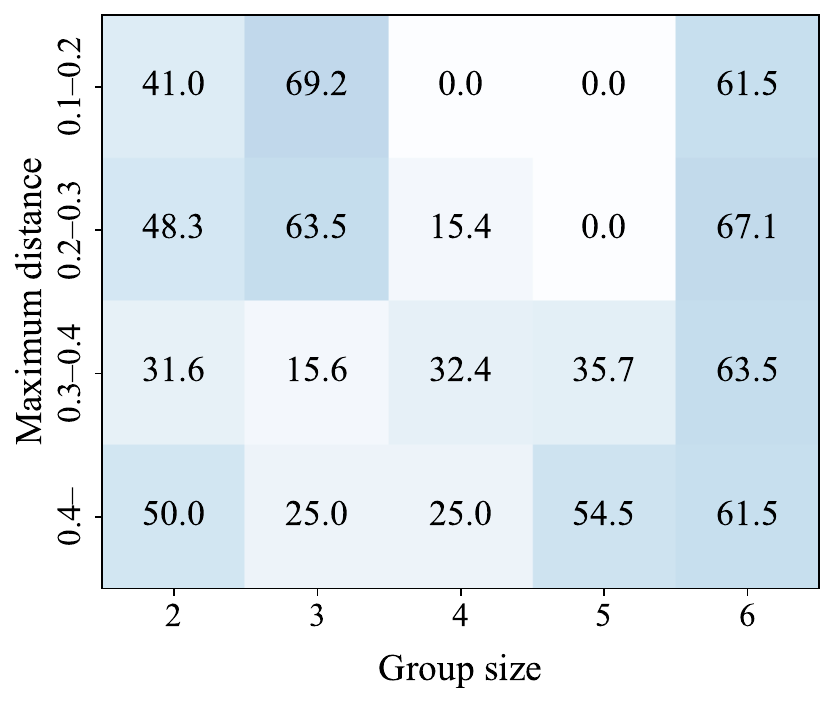}
        \caption{Ours previous~\cite{tamura_eccv2022}}
        \label{fig:dist_size_ours_prev}
    \end{subfigure}
    \hspace{5ex}
    \begin{subfigure}[b]{0.44\linewidth}
        \centering
        \includegraphics[keepaspectratio,width=\textwidth]{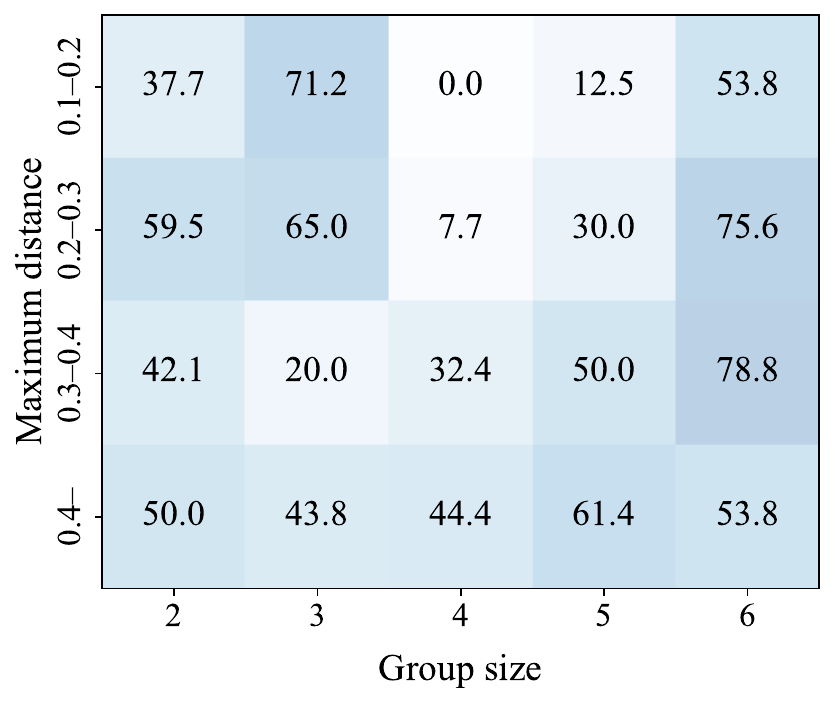}
        \caption{Ours}
        \label{fig:dist_size_ours}
    \end{subfigure}
    \caption{Performances by group sizes and maximum distances of group members. The depth of the color reflects the value in each cell.}
    \label{fig:dist_size}
\end{figure*}

To further analyze the effectiveness of the proposed method in terms of group sizes and distribution of group members in frames, we split the test samples of the Volleyball dataset based on the group sizes and maximum distances of group members and evaluate the performances of Ehsanpour~\etal's method~\cite{ehsanpour_eccv2020}, GroupFormer~\cite{li_iccv2021}, and our methods. Figure.~\ref{fig:dist_size} shows the evaluation results. We also illustrate the number of samples in the training and test set so that the impact of the number can easily be found. Note that the results of samples whose group sizes are one are excluded from the figure to analyze the recognition of genuine social groups.

It can be seen from the figures that the existing two methods have their own preferences for recognition. Figure.~\ref{fig:dist_size_ehsam} shows that Ehsanpour~\etal's method performs well in the case where group members are closely located. Since the method generates individual embeddings with region features, closely located group members can easily be clustered with their similar embeddings, and as a result, the method shows higher performance in that case. GroupFormer, on the other hand, tends to demonstrate better performance when group sizes are large as depicted in Fig.~\ref{fig:dist_size_gf}. In the evaluation of GroupFormer, we set the detection threshold of Deformable DETR to a low value so that it can detect at least one group member in a frame for each sample. The low threshold leads to over-detection and thus degrades the performances of small social groups. The performances of our methods differ from those of the existing methods in that the performances seem to be moderately correlated to the number of training samples rather than indicate specific biases. The results suggest that our methods are highly likely to achieve high performance in recognizing any sort of social group if sufficient training data are provided.

In terms of the performance difference between the proposed and previous methods, the proposed method enhances the performance in cases where group members are far apart or group sizes are large. As described in Sec.~\ref{subsubsec:comp_sgar}, the proposed method has advantages over the previous method due to utilizing multiple embeddings for each social group. The results further clarify the superiority of the proposed method to the previous method.

\subsubsection{Analysis of Individual Recognition Effect}\label{subsubsec:ana_indrec}

\begin{table}[t]
    \caption{Frequent individual actions in the groups of each activity. Note that ``Right" groups are used in this analysis.}
    \label{table:analysis_indact_freq}
    \centering
    \setlength{\tabcolsep}{0.8pt}
    \begin{tabular}{@{}lllllllllll@{}}
        \toprule
        Activity & \multicolumn{5}{c}{Frequent actions} & \multicolumn{5}{c}{Action ratios} \\
        \midrule
        Set & Set & / & Move & / & Stand\hspace{6pt} & \qty{97.9}{\percent} & / & \qty{19.6}{\percent} & / & \qty{1.72}{\percent} \\
        Spike & Spike & / & Block & & & \qty{97.0}{\percent} & / & \qty{75.8}{\percent} && \\
        Pass & Dig & / & Fall &&& \qty{51.4}{\percent} & / & \qty{32.8}{\percent} && \\
        Winpoint\hspace{6pt} & Stand &&&&& \qty{9.12}{\percent} &&&&  \\
        \bottomrule
    \end{tabular}
\end{table}

Since Ehsanpour~\etal's method~\cite{ehsanpour_eccv2020} identifies group members by clustering individual embeddings, which are also used to recognize individual actions, the performance of the method is likely to be affected by individual recognition. To confirm this effect, we first analyze the frequent actions of group members in each activity with the performance of Ehsanpour~\etal's method in Table~\ref{table:comp_socialgroup_volley}. Table~\ref{table:analysis_indact_freq} shows actions performed by more than \qty{20}{\percent} of group members in each activity. The ratios in the table are those of individuals performing the frequent actions in the groups of each activity to total individuals doing the action, which indicate how specific actions are to activities. For instance, most group members in the ``Spike" activity do either the ``Spike" or ``Block" actions, and \qty{75.8}{\percent} of individuals doing the ``Block" action are included in the social groups of the ``Spike" activity. It can be seen from Table~\ref{table:comp_socialgroup_volley} and Table~\ref{table:analysis_indact_freq} that the performances of Ehsanpour~\etal's method and the ratios are positively correlated. Most group members in the ``Spike" activity do either the ``Spike" or ``Block" action, while only a few others do the actions. These activity-specific actions enable backbone networks to embed semantics to identify group members into region features. However, common actions such as ``Stand" complicate the embedding clustering, resulting in the low performances of the ``Set" and ``Winpoint" activities. These results suggest the impact of individual actions on the performance of Ehsanpour~\etal's method.

\begin{table}[t]
    \caption{Comparison of Ehsanpour~\etal's method and the proposed method with and without individual action recognition on social group activity recognition.}
    \label{table:analysis_indact_perf}
    \centering
    \setlength{\tabcolsep}{23pt}
    \begin{threeparttable}
        \begin{tabular}{@{}lcc@{}}
            \toprule
            Method & Action & Accuracy \\
            \midrule
            Ehsanpour~\etal~\cite{ehsanpour_eccv2020}\tnote{\dag} & \checkmark & 44.5 \\
            Ehsanpour~\etal~\cite{ehsanpour_eccv2020}\tnote{\dag} & & 57.6 \\
            \multicolumn{3}{@{}c@{}}{\makebox[\linewidth]{\dashrule[black]}} \\
            Ours & \checkmark & 64.4 \\
            Ours & & 62.1 \\
            \bottomrule
        \end{tabular}
        \begin{tablenotes}\footnotesize
            \item[\dag] Because the source codes are not publicly available, we implemented their algorithm based on our best understanding and evaluated the performance.
        \end{tablenotes}
    \end{threeparttable}
\end{table}

To further analyze the effect of individual action recognition, we evaluate the performances of Ehsanpour~\etal's method and the proposed method without action recognition training. Table~\ref{table:analysis_indact_perf} shows the evaluation results. As seen from the table, the joint training of action recognition drastically degrades the performance of Ehsanpour~\etal's method, while slightly improving the performance of the proposed method. The results of Ehsanpour~\etal's method indicate that action recognition renders the embedding clustering complicated and thus degrades the performance. On the other hand, since the proposed method can selectively aggregate features from feature maps supplemented by action recognition, it does not degrade and rather enhances the performance. The results suggest the susceptibility of Ehsanpour~\etal's method to action recognition and the advantage of the proposed method.

\subsubsection{Analysis of Query Design}\label{subsubsec:ana_qdes_sgar}

\begin{table}[t]
    \caption{Comparison of the naive and decomposed group query implementation on social group activity recognition.}
    \label{table:comp_socialgroup_divquery}
    \centering
    \begin{tabular}{@{}lc@{}}
        \toprule
        Method & Accuracy \\
        \midrule
        Naive query implementation & 63.7 \\
        Decomposed query implementation & 64.4 \\
        \bottomrule
    \end{tabular}
\end{table}

As in the experiments of group activity recognition, we compare the performances of the naive and decomposed group query implementation to analyze the effectiveness. Table~\ref{table:comp_socialgroup_divquery} illustrates the comparison results. As observed in the comparison of group activity recognition in Sec.~\ref{subsubsec:ana_qdes_gar}, the decomposed query implementation slightly improves the performance of the naive query implementation. Since the decomposed query implementation enables group queries to share the knowledge of a group member layout during training, the implementation helps group queries learn the localization of group members. Consequently, the decomposed query implementation achieves better performance than the naive implementation.

\subsubsection{Analysis of Self-Attention Design}\label{subsubsec:ana_attn_sgar}

\begin{table}[t]
    \caption{Comparison of three divided self-attention designs on social group activity recognition. Inter-group and intra-group refer to the attention modules depicted in Fig.~\ref{fig:self_attn_divide}.}
    \label{table:comp_socialgroup_divattndes}
    \centering
    \begin{tabular}{@{}lc@{}}
        \toprule
        Method & Accuracy \\
        \midrule
        Only inter-group & 60.3 \\
        Intra-group $\rightarrow$ inter-group & 63.2 \\
        Inter-group $\rightarrow$ intra-group & 64.4 \\
        \bottomrule
    \end{tabular}
\end{table}

Following the analysis of group activity recognition in Sec.~\ref{subsubsec:ana_attn_gar}, we first compare the three designs of the divided self-attention module on social group activity recognition. Table~\ref{table:comp_socialgroup_divattndes} shows the comparison results. As seen from the table, the design in Fig.~\ref{fig:self_attn_divide} achieves the best performance in the tested designs. Since the group assignment information can be distributed to embeddings in a proper manner with the design in Fig.~\ref{fig:self_attn_divide}, it outperforms the other two variants.

\begin{table}[t]
    \caption{Comparison of self-attention designs on social group activity recognition.}
    \label{table:comp_socialgroup_divattn}
    \centering
    \setlength{\tabcolsep}{12.2pt}
    \begin{tabular}{@{}lccc@{}}
        \toprule
        Method & $N_{id}$ & Divided attn. & Accuracy \\
        \midrule
        Prev.~\cite{tamura_eccv2022} & 1 & & 60.6 \\
        \multicolumn{4}{@{}c@{}}{\makebox[\linewidth]{\dashrule[black]}} \\
        Divided attn. & 12 & \checkmark & 64.4 \\
        Naive attn. & 6 & & 61.6 \\
        \bottomrule
    \end{tabular}
\end{table}

\begin{table}[t]
    \caption{Comparison of self-attention designs on group member identification.}
    \label{table:comp_socialgroup_anagid}
    \centering
    \setlength{\tabcolsep}{4pt}
    \begin{tabular}{@{}lccc@{}}
        \toprule
        Method & $N_{id}$ & Duplicated ratio & Size accuracy \\
        \midrule
        Divided attn. & 12 & 9.94 & 74.2 \\
        Naive attn. & 6 & 11.7 & 72.4 \\
        \bottomrule
    \end{tabular}
\end{table}

We then compare the divided attention of the optimal design with the naive implementation in social group activity recognition. Table~\ref{table:comp_socialgroup_divattn} shows the comparison results. As demonstrated in the table, the divided attention outperforms the naive implementation by a large margin. Since the divided attention demonstrates worse performance on group activity recognition than the naive implementation, the higher performance of the divided attention on social group activity recognition is attributed to the improvement of the group member identification. In the intra-group attention modules of the divided attention, attention values are calculated with embeddings of the same group. This intra-group attention enables embeddings to have detailed communication inside a group and thus helps the assignment of embeddings to group members. As a result, the performance of group member identification is improved. To verify this assumption, we analyze the performances of the naive and divided attention implementations on group member identification. Table~\ref{table:comp_socialgroup_anagid} shows the analysis results. The duplicated ratio indicates the ratio of individuals whose bounding box centers match more than one group member point when minimum distance matching is used instead of Hungarian matching during group member identification. High duplicated ratios mean that multiple embeddings of the same group tend to be assigned to the same individuals. The size accuracy indicates the ratio of samples whose predicted group sizes are correct. As can be seen in the duplicated ratios, the naive implementation is more likely to assign multiple embeddings to the same individuals, which is probably due to the deficient communication of embeddings inside a group. The misassignment renders the prediction of group sizes difficult and thus degrades the performance. These results suggest that the close communication of embeddings inside a group is crucial to identify group members and that the divided attention implementation is more suitable to social group activity recognition than the naive implementation.

\subsubsection{Qualitative Analysis}

\begin{figure*}[!htb]
    \centering
    \begin{subfigure}[b]{0.8\linewidth}
        \centering
        \includegraphics[keepaspectratio,width=\textwidth]{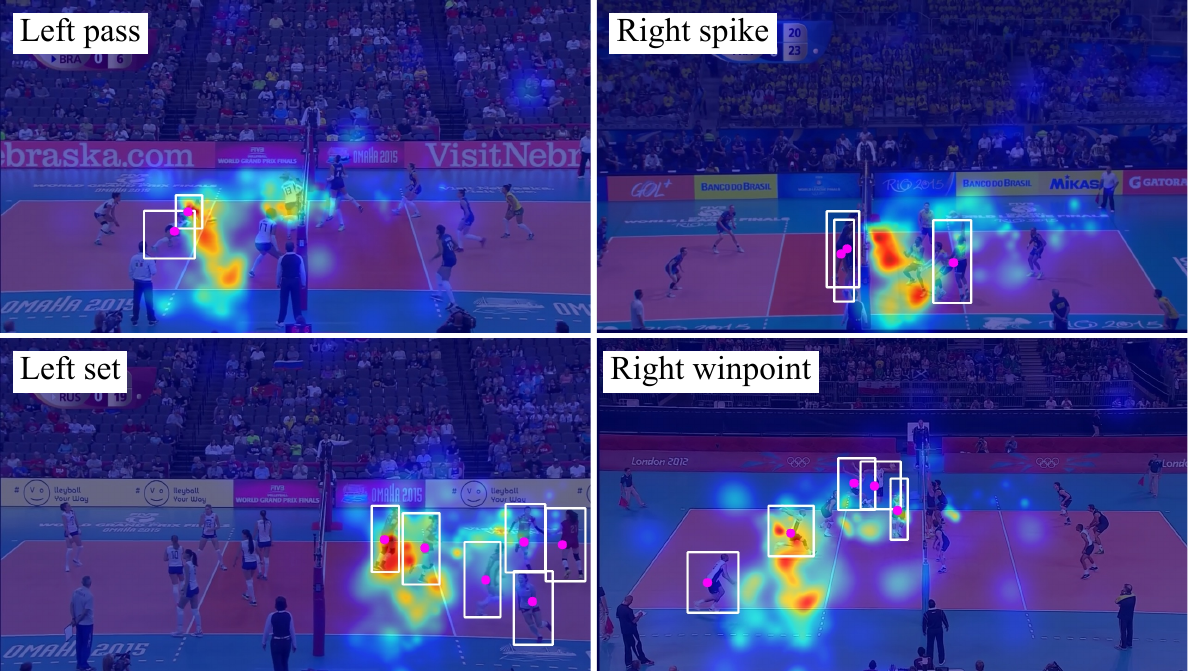}
        \caption{Successful cases.}
        \label{fig:quality_success}
    \end{subfigure} \\
    \begin{subfigure}[b]{0.8\linewidth}
        \centering
        \includegraphics[keepaspectratio,width=\textwidth]{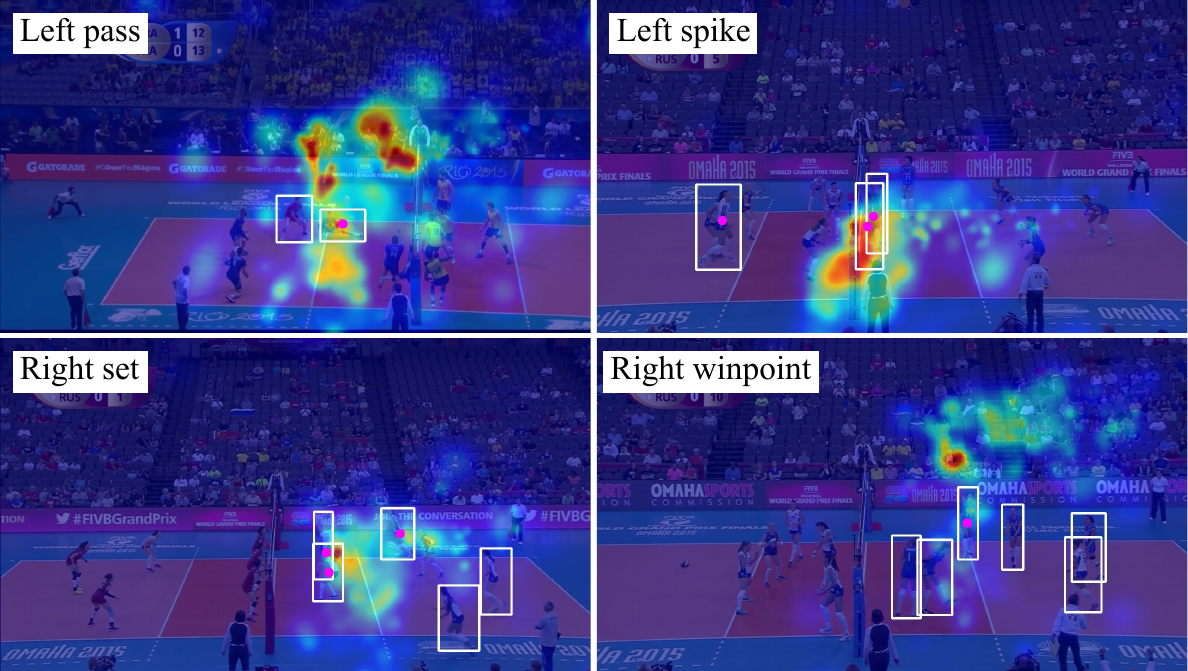}
        \caption{Failure cases.}
        \label{fig:quality_failure}
    \end{subfigure}
    \caption{Visualization of the social group activity recognition results. The white bounding boxes and purple points show the ground truth group members and predicted group member points, respectively. The attention heatmaps are estimated based on the attention locations and weights in the deformable transformer decoder.}
    \label{fig:quality}
\end{figure*}

We analyze the recognition results qualitatively with the success and failure cases of the proposed method. The results of the success and failure cases are shown in Fig.~\ref{fig:quality}. The white bounding boxes and purple points show the ground truth group members and predicted group member points, respectively. The attention heatmaps are estimated based on the attention locations and weights in the deformable transformer decoder, offering a rough range of image areas affecting the embeddings for the predictions.

Although the proposed method misses group members in some cases as depicted in the ``Left spike", ``Right set" and ``Right winpoint" of Fig.~\ref{fig:quality_failure}, the method can recognize social group activities even when group members are apart from each other such as the cases of ``Left set'' and ``Right winpoint'' in Fig.~\ref{fig:quality_success}. The heatmaps show that features are aggregated from the areas around the group members in the successful cases, while in the failure cases, those are aggregated from the wrong areas or areas around only some of the group members. These results indicate that widespread feature aggregation is crucial to recognize social groups when group members are distantly located. Since each embedding can aggregate features from its own target area in feature maps, utilizing multiple embeddings for a group is reasonable for widespread feature aggregation.

\section{Conclusion}\label{sec:conc}

In this paper, we propose and analyze efficient designs of group queries and self-attention implementations in transformers for social group activity recognition. Our previous method achieves state-of-the-art performance on social group activity recognition by leveraging transforms to generate effective social group features. However, the method has limited capability to recognize social groups of large sizes due to a complicated encoding of group member locations. To solve this issue, we propose utilizing multiple query embeddings as a group query to recognize social group activities. We analyze the efficient ways of creating group queries and calculating attention values in a transformer decoder because the increased number of embeddings renders group query training difficult and communication of embeddings deficient. Experimental results demonstrate that the proposed method outperforms existing and our previous methods and enhances the performance especially when groups have large sizes or distantly located members.

\begin{acknowledgements}
Computational resource of AI Bridging Cloud Infrastructure (ABCI) provided by National Institute of Advanced Industrial Science and Technology (AIST) was used.
\end{acknowledgements}

\bibliographystyle{spmpsci}
\bibliography{ref}

\begin{thebibliography}{10}
\providecommand{\url}[1]{{#1}}
\providecommand{\urlprefix}{URL }
\expandafter\ifx\csname urlstyle\endcsname\relax
  \providecommand{\doi}[1]{DOI~\discretionary{}{}{}#1}\else
  \providecommand{\doi}{DOI~\discretionary{}{}{}\begingroup \urlstyle{rm}\Url}\fi

\bibitem{amer_eccv2014}
Amer, M.R., Lei, P., Todorovic, S.: {HiRF}: {H}ierarchical random field for collective activity recognition in videos.
\newblock In: ECCV (2014)

\bibitem{amer_tpami2016}
Amer, M.R., Todorovic, S.: Sum product networks for activity recognition.
\newblock IEEE TPAMI \textbf{38}(4), 800--813 (2016)

\bibitem{amer_iccv2013}
Amer, M.R., Todorovic, S., Fern, A., Zhu, S.C.: Monte carlo tree search for scheduling activity recognition.
\newblock In: ICCV (2013)

\bibitem{azar_cvpr2019}
Azar, S.M., Atigh, M.G., Nickabadi, A., Alahi, A.: Convolutional relational machine for group activity recognition.
\newblock In: CVPR (2019)

\bibitem{bagautdinov_cvpr2017}
Bagautdinov, T.M., Alahi, A., Fleuret, F., Fua, P.V., Savarese, S.: Social scene understanding: {E}nd-to-end multi-person action localization and collective activity recognition.
\newblock In: CVPR (2017)

\bibitem{bertasius_icml2021}
Bertasius, G., Wang, H., Torresani, L.: Is space-time attention all you need for video understanding?
\newblock In: ICML (2021)

\bibitem{carion_eccv2020}
Carion, N., Massa, F., Synnaeve, G., Usunier, N., Kirillov, A., Zagoruyko, S.: End-to-end object detection with transformers.
\newblock In: ECCV (2020)

\bibitem{carreira_cvpr2017}
Carreira, J., Zisserman, A.: Quo vadis, action recognition? {A} new model and the kinetics dataset.
\newblock In: CVPR (2017)

\bibitem{choi_eccv2014}
Choi, W., Chao, Y.W., Pantofaru, C., Savarese, S.: Discovering groups of people in images.
\newblock In: ECCV (2014)

\bibitem{choi_iccvw2009}
Choi, W., Shahid, K., Savarese, S.: What are they doing? : {C}ollective activity classification using spatio-temporal relationship among people.
\newblock In: ICCVW (2009)

\bibitem{dai_iccv2017}
Dai, J., Qi, H., Xiong, Y., Li, Y., Zhang, G., Hu, H., Wei, Y.: Deformable convolutional networks.
\newblock In: ICCV (2017)

\bibitem{deng_cvpr2016}
Deng, Z., Vahdat, A., Hu, H., Mori, G.: Structure inference machines: Recurrent neural networks for analyzing relations in group activity recognition.
\newblock In: CVPR (2016)

\bibitem{ehsanpour_eccv2020}
Ehsanpour, M., Abedin, A., Saleh, F., Shi, J., Reid, I., Rezatofighi, H.: Joint learning of social groups, individuals action and sub-group activities in videos.
\newblock In: ECCV (2020)

\bibitem{gavrilyuk_cvpr2020}
Gavrilyuk, K., Sanford, R., Javan, M., Snoek, C.G.M.: Actor-transformers for group activity recognition.
\newblock In: CVPR (2020)

\bibitem{weina_tpami2012}
Ge, W., Collins, R.T., Ruback, R.B.: Vision-based analysis of small groups in pedestrian crowds.
\newblock IEEE TPAMI \textbf{34}(5), 1003--1016 (2012)

\bibitem{hu_cvpr2020}
Hu, G., Cui, B., He, Y., Yu, S.: Progressive relation learning for group activity recognition.
\newblock In: CVPR (2020)

\bibitem{ibrahim_eccv2018}
Ibrahim, M.S., Mori, G.: Hierarchical relational networks for group activity recognition and retrieval.
\newblock In: ECCV (2018)

\bibitem{ibrahim_cvpr2016}
Ibrahim, M.S., Muralidharan, S., Deng, Z., Vahdat, A., Mori, G.: A hierarchical deep temporal model for group activity recognition.
\newblock In: CVPR (2016)

\bibitem{kay_arxiv2017}
Kay, W., Carreira, J., Simonyan, K., Zhang, B., Hillier, C., Vijayanarasimhan, S., Viola, F., Green, T., Back, T., Natsev, P., Suleyman, M., Zisserman, A.: The kinetics human action video dataset (2017).
\newblock ArXiv:1705.06950

\bibitem{kipf_iclr2017}
Kipf, T.N., Welling, M.: Semi-supervised classification with graph convolutional networks.
\newblock In: ICLR (2017)

\bibitem{kong_icassp2018}
Kong, L., Qin, J., Huang, D., Wang, Y., Gool, L.V.: Hierarchical attention and context modeling for group activity recognition.
\newblock In: ICASSP (2018)

\bibitem{kuhn_naval1955}
Kuhn, H.W., Yaw, B.: The hungarian method for the assignment problem.
\newblock Naval Res. Logist. Quart pp. 83--97 (1955)

\bibitem{lan_cvpr2012}
Lan, T., Sigal, L., Mori, G.: Social roles in hierarchical models for human activity recognition.
\newblock In: CVPR (2012)

\bibitem{lan_tpami2012}
Lan, T., Wang, Y., Yang, W., Robinovitch, S.N., Mori, G.: Discriminative latent models for recognizing contextual group activities.
\newblock IEEE TPAMI \textbf{34}(8), 1549--1562 (2012)

\bibitem{li_iccv2021}
Li, S., Cao, Q., Liu, L., Yang, K., Liu, S., Hou, J., Yi, S.: Group{F}ormer: {G}roup activity recognition with clustered spatial-temporal transformer.
\newblock In: ICCV (2021)

\bibitem{li_iccv2017}
Li, X., Chuah, M.C.: {SBGAR}: {S}emantics based group activity recognition.
\newblock In: ICCV (2017)

\bibitem{lin_iccv2017}
Lin, T.Y., Goyal, P., Girshick, R., He, K., Doll{\'a}r, P.: Focal loss for dense object detection.
\newblock In: ICCV (2017)

\bibitem{lin_eccv2014}
Lin, T.Y., Maire, M., Belongie, S., Hays, J., Perona, P., Ramanan, D., Doll{\'a}r, P., Zitnick, C.L.: Microsoft {COCO}: {C}ommon objects in context.
\newblock In: ECCV (2014)

\bibitem{loshchiloy_iclr2019}
Loshchilov, I., Hutter, F.: Decoupled weight decay regularization.
\newblock In: ICLR (2019)

\bibitem{park_cvpr2015}
Park, H., Shi, J.: Social saliency prediction.
\newblock In: CVPR (2015)

\bibitem{pramono_eccv2020}
Pramono, R.R.A., Chen, Y.T., Fang, W.H.: Empowering relational network by self-attention augmented conditional random fields for group activity recognition.
\newblock In: ECCV (2020)

\bibitem{qi_eccv2018}
Qi, M., Qin, J., Li, A., Wang, Y., Luo, J., Gool, L.V.: stag{N}et: {A}n attentive semantic rnn for group activity recognition.
\newblock In: ECCV (2018)

\bibitem{sendo_mva2019}
Sendo, K., Ukita, N.: Heatmapping of people involved in group activities.
\newblock In: MVA (2019)

\bibitem{shu_cvpr2017}
Shu, T., Todorovic, S., Zhu, S.C.: {CERN}: {C}onfidence-energy recurrent network for group activity recognition.
\newblock In: CVPR (2017)

\bibitem{tamura_eccv2022}
Tamura, M., Vishwakarma, R., Vennelakanti, R.: Hunting group clues with transformers for social group activity recognition.
\newblock In: ECCV (2022)

\bibitem{jinhui_tpami2022}
Tang, J., Shu, X., Yan, R., Zhang, L.: Coherence constrained graph lstm for group activity recognition.
\newblock IEEE TPAMI \textbf{44}(2), 636--647 (2022)

\bibitem{vaswani_nips2017}
Vaswani, A., Shazeer, N., Parmar, N., Uszkoreit, J., Jones, L., Gomez, A.N., Kaiser, L., Polosukhin, I.: Attention is all you need.
\newblock In: NIPS (2017)

\bibitem{velickovic_iclr2018}
Veli{\v{c}}kovi{\v{c}}, P., Cucurull, G., Casanova, A., Romero, A., Liò, P., Bengio, Y.: Graph attention networks.
\newblock In: ICLR (2018)

\bibitem{wang_cvpr2017}
Wang, M., Ni, B., Yang, X.: Recurrent modeling of interaction context for collective activity recognition.
\newblock In: CVPR (2017)

\bibitem{wang_cvpr2013}
Wang, Z., Shi, Q., Shen, C., van~den Hengel, A.: Bilinear programming for human activity recognition with unknown mrf graphs.
\newblock In: CVPR (2013)

\bibitem{wu_cvpr2019}
Wu, J., Wang, L., Wang, L., Guo, J., Wu, G.: Learning actor relation graphs for group activity recognition.
\newblock In: CVPR (2019)

\bibitem{yan_tnnls2021}
Yan, R., Shu, X., Yuan, C., Tian, Q., Tang, J.: Position-aware participation-contributed temporal dynamic model for group activity recognition.
\newblock IEEE TNNLS \textbf{33}(12), 7574--7588 (2022)

\bibitem{yan_tpami2020}
Yan, R., Xie, L., Tang, J., Shu, X., Tian, Q.: {HiGCIN}: {H}ierarchical graph-based cross inference network for group activity recognition.
\newblock IEEE TPAMI  (2020)

\bibitem{yan_eccv2020}
Yan, R., Xie, L., Tang, J., Shu, X., Tian, Q.: Social adaptive module for weakly-supervised group activity recognition.
\newblock In: ECCV (2020)

\bibitem{yuan_iccv2021}
Yuan, H., Ni, D., Wang, M.: Spatio-temporal dynamic inference network for group activity recognition.
\newblock In: ICCV (2021)

\bibitem{zhou_arxiv2021}
Zhou, H., Kadav, A., Shamsian, A., Geng, S., Lai, F., Zhao, L., Liu, T., Kapadia, M., Graf, H.P.: {COMPOSER}: {C}ompositional learning of group activity in videos.
\newblock arXiv preprint arXiv:2112.05892  (2021)

\bibitem{zhou_arxiv2019}
Zhou, X., Wang, D., Kr{\"a}henb{\"u}hl, P.: Objects as points (2019).
\newblock ArXiv:1904.07850

\bibitem{zhu_iclr2021}
Zhu, X., Su, W., Lu, L., Li, B., Wang, X., Dai, J.: Deformable {DETR}: {D}eformable transformers for end-to-end object detection.
\newblock In: ICLR (2021)

\end{thebibliography}

\end{document}